\documentclass[sigconf,natbib,screen=true]{acmart}

\PassOptionsToPackage{dvipsnames}{xcolor}

\usepackage{dsfont}
\usepackage{acronym}
\usepackage{algorithm}
\usepackage[noend]{algpseudocode}
\usepackage{bbm}
\usepackage{beramono}
\usepackage{bm}
\usepackage{booktabs}
\usepackage[skip=0pt]{caption}
\usepackage{cleveref}
\usepackage[T1]{fontenc}
\usepackage{graphicx}
\usepackage{hyperref}
\usepackage{listings}
\usepackage{multirow}
\usepackage{natbib}
\usepackage{subcaption}
\usepackage{tabularx}
\usepackage[dvipsnames]{xcolor}
\usepackage{enumerate}
\usepackage[inline]{enumitem}
\usepackage{numprint}
\usepackage[normalem]{ulem}

\copyrightyear{2021}
\acmYear{2021}
\setcopyright{iw3c2w3}
\acmConference[WWW '21]{Proceedings of the Web Conference 2021}{April 19--23, 2021}{Ljubljana, Slovenia}
\acmBooktitle{Proceedings of the Web Conference 2021 (WWW '21), April 19--23, 2021, Ljubljana, Slovenia}

\acmPrice{}
\acmDOI{10.1145/3442381.3450018}
\acmISBN{978-1-4503-8312-7/21/04}

\definecolor{todored}{RGB}{150, 0, 0}
\definecolor{rj}{RGB}{0, 150, 0}
\definecolor{mdr}{RGB}{200, 0, 0}
\definecolor{ho}{RGB}{0, 50, 150}

\acrodef{IR}{Information Retrieval}
\acrodef{LTR}{Learning to Rank}
\acrodef{ARP}{Average Relevance Position}
\acrodef{DCG}{Discounted Cumulative Gain}
\acrodef{EM}{Expectation Maximization}
\acrodef{GENSPEC}{Generalization and Specialization}
\acrodef{IPS}{Inverse Propensity Scoring}
\acrodef{CTR}{Click-Through-Rate}
\acrodef{SEA}{Safe Exploration Algorithm}
\acrodef{PBM}{Position-Based Model algorithm}

\DeclareMathOperator*{\argmax}{arg\,max}

\newcommand{\data}{\mathcal{D}}
\newcommand{\traindata}{\mathcal{D}^\textit{train}}
\newcommand{\bounddata}{\mathcal{D}^\textit{sel}}

\allowdisplaybreaks

\author{Harrie Oosterhuis}
\affiliation{%
	\institution{Radboud University}
	\city{Nijmegen}
	\country{The Netherlands}
}
\email{harrie.oosterhuis@ru.nl}
 
\author{Maarten de Rijke}
\orcid{0000-0002-1086-0202}
\affiliation{
 \institution{University of Amsterdam \& Ahold Delhaize}
 \city{Amsterdam}
 \country{The Netherlands}
}
\email{derijke@uva.nl}

\acmSubmissionID{1559}

\title[Robust Generalization and Safe Query-Specialization in Counterfactual Learning to Rank]{Robust Generalization and Safe Query-Special\-ization\\in Counterfactual Learning to Rank}

\begin{document}

\begin{abstract}

Existing work in counterfactual \ac{LTR} has focussed on optimizing feature-based models that predict the optimal ranking based on document features.
\ac{LTR} methods based on bandit algorithms often optimize tabular models that memorize the optimal ranking per query.
These types of model have their own advantages and disadvantages.
Feature-based models provide very robust performance across many queries, including those previously unseen, however, the available features often limit the rankings the model can predict.
In contrast, tabular models can converge on any possible ranking through memorization.
However, memorization is extremely prone to noise, which makes tabular models reliable only when large numbers of user interactions are available.
Can we develop a robust counterfactual \ac{LTR} method that pursues memorization-based optimization whenever it is safe to do?

We introduce the \acf{GENSPEC} algorithm, a robust feature-based counterfactual \ac{LTR} method that pursues per-query memorization when it is safe to do so. \ac{GENSPEC} optimizes a single feature-based model for generalization: robust performance across all queries, and many tabular models for specialization: each optimized for high performance on a single query.
\ac{GENSPEC} uses novel relative high-confidence bounds to choose which model to deploy per query.
By doing so, \ac{GENSPEC} enjoys the high performance of successfully specialized tabular models with the robustness of a generalized feature-based model. 
Our results show that \ac{GENSPEC} leads to optimal performance on queries with sufficient click data, while having robust behavior on queries with little or noisy data.
\end{abstract}

\maketitle

\acresetall

\section{Introduction}
\label{sec:intro}

Ranking systems form the basis for search and recommendation services on the world wide web~\citep{sanderson2010user}. %
The field of \ac{LTR} considers methods that optimize ranking systems~\citep{liu2009learning, jarvelin2002cumulated, schafer1999recommender}.
An important branch of \ac{LTR} research is based on learning from user interactions~\citep{jagerman2019comparison, joachims2017unbiased, hofmann2013reusing, Joachims2002}. This approach has several advantages.
For a service with an active user base, user interactions are virtually free and widely available~\citep{joachims2017unbiased}.
Interactions allow methods to closely learn user preferences~\citep{sanderson2010user}, even in cases where expert annotations cannot be obtained~\citep{wang2016learning}.
However, user interactions are affected by noise and bias.
Interactions on rankings are particularly affected by \emph{position bias}~\citep{craswell2008experimental}:
items often receive more clicks due to their display-position, and not due to being more preferred by users.
Therefore \ac{LTR} methods that learn from user interactions have to correct for the forms of bias that affect them~\citep{joachims2017unbiased}.

In previous work, these \ac{LTR} methods have been divided into online and counterfactual approaches~\citep{jagerman2019comparison, ai2020unbiased, jagerman2020safety}, where online approaches learn from direct interactions~\citep{yue2009interactively, oosterhuis2018differentiable, zhuang2020counterfactual}, and counterfactual approaches learn from historical interaction data~\citep{wang2016learning, joachims2017unbiased, oosterhuis2020topkrankings}.
While this division is very interesting~\citep{oosterhuis2021onlinecounterltr, ai2020unbiased, jagerman2019comparison}, this paper focusses on a different division between methods that learn feature-based models and those that learn tabular models.
The former group of methods optimize models that predict the optimal ranking based on the available document features, this group includes most work on counterfactual \ac{LTR}~\citep{wang2016learning, joachims2017unbiased, oosterhuis2020topkrankings} and online \ac{LTR} methods such as Dueling Bandit Gradient Descent~\citep{yue2009interactively}, Pairwise Differentiable Gradient Descent~\citep{oosterhuis2018differentiable} and the Counterfactual Online Learning to Rank algorithm~\citep{zhuang2020counterfactual}.
The latter group of methods do not predict based on document features, instead they attempt to memorize the optimal ranking.
Thus they try to optimize the ranking directly, this group includes methods such as \ac{PBM}~\citep{lagree2016multiple}, the Hotfix algorithm~\citep{zoghi2016click}, and various approaches based on $k$-armed bandit algorithms~\citep{katariya2016dcm, kveton2015cascading, Radlinski2008, li2020bubblerank, lattimore2019bandit}.

Feature-based models are very good at \emph{generalization}, they perform well across large groups of queries including rare or previously unseen ones~\citep{liu2009learning}.
However, feature-based models usually do not reach optimal performance because they are limited by the available features~\citep{zoghi2016click}.
For instance, often the features do not contain enough information to predict the optimal ranking. %
In contrast, tabular models are well-suited for \emph{specialization}, they can reach optimal performance on individual queries where enough interactions are available~\citep{lagree2016multiple}.
Because they do not utilize feature-based predictions, tabular models can converge on any possible ranking, and thus provide extremely high performance if enough interaction data is available~\citep{zoghi2016click}.
Unfortunately, tabular models can also provide extremely poor performance when not enough data has been gathered.
This happens because the ranking behavior memorized on one query cannot be generalized to other queries~\citep{zoghi2016click}.
For instance, tabular models have no preference between rankings on previously unseen queries.
Moreover, because their behavior does not generalize, tabular models are very sensitive to noise on infrequent queries where little data is available.

Therefore, the important choice between feature-based models and tabular models  should mainly depend on the amount of available data per query.
Feature-based models are the best choice when little or no interactions on a query are available, while tabular models are the better choice on queries where large numbers of interactions have been gathered.
This observation has been made in previous work, i.e., \citet{zoghi2016click} noted that their Hotfix algorithm should only be applied to underperforming \emph{torso-queries}, and explicitly advised not to apply the algorithm to infrequent \emph{tail} queries.
In practice, tabular models are applied to queries for which feature-based ranking systems appear to be underperforming and that receive enough interactions so that improvements can be found~\citep{trotman-2017-architecture,mavridis-2020-beyond,grainger-2020-ai}.
To the best of our knowledge, there is no theory-grounded approach for deciding when it is safe to pursue optimizations offered by a tabular model over the robustness offered by a feature-based model, despite the fact that these choices are being made on a daily basis in real-world production systems.

Inspired by the advantages of both types of models and the lack of a principled approach to choosing and switching between them, we introduce the \acfi{GENSPEC}\acused{GENSPEC} framework.
Using counterfactual \ac{LTR}, \ac{GENSPEC} optimizes multiple ranking models for either generalization across queries or specialization in a specific query.
Using the available click data \ac{GENSPEC} simultaneously trains:
\begin{enumerate*}[label=(\roman*)]
\item a generalized feature-based model that performs well across all queries, and 
\item a specialized tabular model that memorizes a ranking for each observed query.
\end{enumerate*}
Per individual query there is the choice between three models:
\begin{enumerate*}[label=(\roman*)]
\item the logging policy model used to gather the click data, 
\item the generalized feature-based model, and 
\item the specialized tabular model.
\end{enumerate*}
To reliably choose between these models, we introduce novel high-confidence bounds on the relative performance between models, based on existing bounds on absolute performance~\citep{thomas2015high}.
Using these novel bounds, \ac{GENSPEC} chooses conservatively:
the feature-based model is only deployed when there is high confidence that it outperforms the logging policy;
and a tabular model overrules the other models on a single query, if with high confidence it is expected to outperform both for that specific query.
By simultaneously deploying both feature-based and tabular models, \ac{GENSPEC} exploits the advantageous properties of both:
the robustness of a generalized feature-based model with the potential high performance of a specialized tabular model.

The main contributions of this work are:
\begin{itemize}[leftmargin=*,nosep]
\item the \ac{GENSPEC} framework that simultaneously optimizes generalized and specialized models and decides which to deploy per individual query; and
\item a novel high-confidence bound for estimating the relative performance difference between ranking models. 
\end{itemize}
To the best of our knowledge, \ac{GENSPEC} is the first framework to simultaneously optimize both feature-based and tabular models based on user interactions, and reliably choose between them on a query-level using high-confidence bounds.

\section{Related Work}
\label{sec:relatedwork}

\ac{LTR} or the optimization of ranking models w.r.t.\ ranking metrics, is a well-established field within \ac{IR}~\citep{liu2009learning}.
Supervised \ac{LTR} methods make use of annotated datasets based on expert judgements.
These datasets contain labels indicating the expert-judged level of relevance for numerous query-document pairs~\citep{Chapelle2011, qin2013introducing, dato2016fast}.
With such an annotated dataset evaluation and optimization is mostly straightforward.
Since ranking metrics are not differentiable, \ac{LTR} methods either optimize an approximate metric or a lower bound on the metric~\citep{liu2009learning, wang2018lambdaloss, agarwal2019counterfactual}.

While supervised \ac{LTR} has a very important place in the \ac{IR} field, several limitations of the supervised approach have become apparent.
Most importantly, expert annotations often disagree with actual user preferences~\citep{sanderson2010user}.
Furthermore, they are expensive and time-consuming to gather~\citep{Chapelle2011, qin2013introducing}.
In privacy-sensitive settings it is often impossible to acquire judgements without breaching the privacy of users, i.e., in search through emails or personal documents~\citep{wang2016learning}.
Consequently, interest in \ac{LTR} from user interactions has increased rapidly in recent years~\citep{wang2016learning, joachims2017unbiased, oosterhuis2020topkrankings, agarwal2019addressing, vardasbi2020trust}.
In contrast to annotated datasets, interactions are cheap and easy to obtain if a service has active users, and can be gathered without showing privacy-sensitive content to experts~\citep{wang2016learning, joachims2017unbiased}.
Moreover, interactions are indicative of actual user preferences~\citep{radlinski2008does}.
Unfortunately, unlike annotations, interactions are noisy and biased indicators of relevance and thus bring their own difficulties~\citep{craswell2008experimental}.

The idea of \ac{LTR} from user clicks goes back to one of the earliest pairwise \ac{LTR} methods~ \citep{Joachims2002}.
More recent work has introduced the field of counterfactual \ac{LTR}~\citep{wang2016learning, joachims2017unbiased}, where \ac{IPS} is used to counter the effect of position bias~\citep{craswell2008experimental}.
The underlying idea is that a model of position bias can be inferred from user interactions reliably, and with such a model \ac{IPS} can correct for its effect~\citep{wang2018position} (see Section~\ref{sec:background}).
Work on counterfactual \ac{LTR} is concentrated around methods for estimating models of bias~\citep{wang2018position, agarwal2019estimating}, and counterfactual estimators that use these models for unbiased evaluation and optimization~\citep{wang2016learning, joachims2017unbiased, agarwal2019counterfactual}.
Recent work has extended the counterfactual \ac{LTR} approach to also correct for item-selection bias~\citep{oosterhuis2020topkrankings} and trust bias~\citep{agarwal2019addressing, vardasbi2020trust}.
Existing work on counterfactual \ac{LTR} has only optimized feature-based models, e.g., support vector machines~\citep{joachims2017unbiased}, linear ranking models~\citep{oosterhuis2020topkrankings}, and neural networks~\citep{agarwal2019counterfactual}.

In contrast, methods that optimize tabular models are common in online \ac{LTR} work~\citep{katariya2016dcm, kveton2015cascading, Radlinski2008, li2020bubblerank, lattimore2019bandit}.
These methods learn from direct interactions with the user where they can decide which rankings will be displayed.
\citet{zoghi2016click} have argued that the main advantage of tabular models is that they are not limited by the available features.
Consequently, they can converge on any possible ranking, allowing for extremely high performance if successfully optimized.
However, tabular models do not generalize across queries, and as a result, these methods see each query as an independent ranking problem.
For this reason, these algorithms have a cold-start problem: they often have an initial period of very poor performance.
To mitigate this, various heuristics (e.g., in terms of business rules, query frequency, or sampling strategies for training) are put in place when using tabular models in practice~\cite{zoghi2016click,sorokina-2016-amazon,grainger-2020-ai}.

Prior work on query frequency has found that they follow a long-tail distribution~\citep{silverstein1999analysis, spink2002us}.
\citet{white2007studying} found that 97\% of queries received fewer than $10$ clicks over six months.
For such infrequent queries, the performance of a tabular model may never leave the initial period of poor performance~\citep{wu2016conservative}.
Even though it is known that deployment should be avoided in such cases, to the best of our knowledge, there exist no theoretically principled method for detecting when it is safe to deploy a tabular model.
The only existing method that safely chooses between models appears to be the \ac{SEA}~\citep{jagerman2020safety}, which applies high-confidence bounds to the performance of a safe logging policy model and a newly learned ranking model.
If these bounds do not overlap, \ac{SEA} can conclude with high-confidence that one model outperforms the other.
So far, no work has used \ac{SEA} for the deployment of tabular models in \ac{LTR}.

\section{Background}
\label{sec:background}

\subsection{The Learning to Rank Task}

The goal of \ac{LTR} is to optimize a ranking model w.r.t.\ to a ranking metric.
We use $y$ to denote a ranking, which is simply an ordering of documents $d$:
$
y = [d_1, d_2, \ldots].
$
For a ranking model, we use $\pi$ and $\pi(y \mid q)$ for the probability of $\pi$ displaying $y$ for query $q$:
\begin{equation}
\pi(y \mid q) = P(Y = y \mid q, \pi).
\end{equation}
Queries follow a distribution determined by the users, we use $P(Q = q)$ to denote the probability that a user-issued query is $q$.
Furthermore, we use $r(q, d)$ to denote the relevance of $d$ w.r.t.\ $q$, this is the probability that a user considers $d$ a relevant result for query $q$:
$
r(q, d) = P(R = 1 \mid q, d).
$
Most ranking metrics compute the quality of a single ranking as a weighted sum over the ranks~\cite{liu2009learning, joachims2017unbiased}.
Let $\textit{rank}(d \mid y) \in \mathds{Z}_{>0}$ indicate the rank of $d$ in $y$ and $\lambda: \mathds{Z}_{>0} \rightarrow \mathds{R}$ be an arbitrary weight function, then
we use $\Delta(y \mid q, r)$ to denote the quality of $y$ w.r.t.\  the query $q$ and the relevance function $r$:
\begin{equation}
\Delta(y \mid q, r) = \sum_{d \in y}\lambda\big(\textit{rank}(d \mid y)\big) \cdot r(q, d).
\end{equation}
Ranking metrics differ in their choice of $\lambda$ and thus they differ in how important they consider each rank to be.
A common ranking metric is the \acfi{DCG} metric~\citep{jarvelin2002cumulated}; $\lambda$ can be chosen accordingly:
\begin{equation}
\lambda^{\textit{DCG}}\big(\textit{rank}(d \mid y)\big) = \log_2\big(1+ \textit{rank}(d \mid y)\big)^{-1}. \label{eq:dcg}
\end{equation}
Finally, the performance of a ranking model $\pi$ according to a ranking metric is an expectation over the query distribution $P(Q = q)$ and the model behavior, that is, the expected ranking quality $\Delta(y \mid q, r)$ w.r.t. the model $\pi$ and the query distribution:
\begin{equation}
\mathcal{R}(\pi) =  \int \Big( \sum_{y \in \pi} \Delta(y \mid q, r) \cdot \pi(y \mid q) \Big) P(Q = q) \,dq.
\label{eq:truereward}
\end{equation}
Supervised \ac{LTR} methods use datasets where often $P(Q = q)$ is approximated based on logged user-issued queries, and the relevance function $r(q,d)$ is estimated using expert judgements.
Given such a dataset, these methods can optimize the resulting estimate of $\mathcal{R}(\pi)$ in a supervised manner~\cite{liu2009learning, burges2010ranknet, wang2018lambdaloss}.

\subsection{Counterfactual Learning to Rank}
\label{sec:counterfactualltr}

As discussed in Section~\ref{sec:relatedwork}, there are severe limitations to estimates of the relevance function $r(q,d)$ based on expert judgements.
Acquiring relevance annotations is expensive, time-consuming~\citep{Chapelle2011, qin2013introducing}, and sometimes infeasible, for instance, due to privacy concerns~\citep{wang2016learning}.
Moreover, the resulting annotations are often not aligned with the actual user preferences~\citep{sanderson2010user}.

An attractive alternative to the approach of collecting expert generated annotations is counterfactual \ac{LTR}, which learns from historical interaction logs~\citep{wang2016learning,joachims2017unbiased}.
Let $\pi_0$ be the logging policy that was used for gathering interactions, with $q_i$ as the user-issued query, $y_i$ as the ranking displayed at interaction $i$ and:
$
y_i \sim \pi_0(y \mid q_i).
$
To model position bias, we will use the $o_i(d) \in \{0,1\}$ to indicate whether item $d$ was examined by the user or not:
$
o_i(d) \sim P\big(O \mid d, y_i\big).
$
Furthermore, we use $c_i(d) \in \{0,1\}$ to indicate whether $d$ was clicked at time step $i$:
$
c_i(d) \sim P\big(C \mid o_i(d), r( d \mid q)\big).
$
We follow the common assumption that users do not click on unexamined documents:
\begin{equation}
P\big(C = 1 \mid o_i(d) = 0, r(q, d)\big) = 0.
\end{equation}
Given that a document is observed, its click probability is assumed to be proportional to its relevance with some constant offset $\mu \in \mathbb{R}_{>0}$:
\begin{equation}
\label{eq:observed}
P\big(C = 1 \mid o_i(d) = 1, r(q, d)\big) \propto r(q, d) + \mu.
\end{equation}
The data used in counterfactual \ac{LTR} consists of the observed clicks $c$, the displayed rankings $y$, the propensity scores $\rho$, and the issued queries $q$ for $N$ interactions:
\begin{equation}
\mathcal{D} = \big\{(c_i, y_i, \rho_i, q_i)\big\}_{i=1}^N.
\end{equation}
We apply the policy-aware approach~\citep{oosterhuis2020topkrankings} and base the propensity scores both on the logging policy $\pi_0$ as the position bias of the user:
\begin{equation}
\rho_i(d) = \sum_{y \in \pi_0} P\big(O=1 \mid d, y_i\big) \cdot \pi_0(y_i \mid q_i).
\end{equation}
The main difficulty in learning from $\mathcal{D}$ is that the observed clicks $c$ are affected by both relevance, the logging policy behavior, and the users' position bias.
Counterfactual \ac{LTR} methods apply \ac{IPS} estimators to correct the click signal for the position bias and logging policy behavior.
The resulting estimated reward can then be used to unbiasedly optimize a ranking model.

The estimated reward is an average over an \ac{IPS} transformation of each point of interaction data:
\begin{equation}
\hat{\mathcal{R}}(\pi \mid \mathcal{D}) = \frac{1}{|\mathcal{D}|} \sum_{i \in \mathcal{D}} \sum_{y \in \pi}  \hat{\Delta}(y \mid c_i, \rho_i) \cdot \pi(y \mid q_i),
\label{eq:rewardestimate}
\end{equation}
where $\hat{\Delta}$ is an \ac{IPS} estimator:
\begin{equation}
\hat{\Delta}(y \mid c_i, \rho_i) = \sum_{d \in y}\lambda\big(\textit{rank}(d \mid y)\big) \cdot  \frac{c_i(d)}{\rho_i(d)}. \label{eq:rankips}
\end{equation}
The estimated reward
$\hat{\mathcal{R}}(\pi \mid \mathcal{D})$
is proven to enable unbiased \ac{LTR} because it maintains the ordering of ranking models (see the proof in Appendix~\ref{sec:counterfactualproof}).
In other words, for any two ranking models $\pi_1, \pi_2$ the estimated reward $\hat{\mathcal{R}}$ will prefer the same model as the true reward $\mathcal{R}$:
\begin{equation}
\forall \pi_1, \pi_2, \,
\mathds{E}\big[\hat{\mathcal{R}}(\pi_1 \mid \mathcal{D}) > \hat{\mathcal{R}}(\pi_2 \mid \mathcal{D})\big]
\leftrightarrow
\mathcal{R}(\pi_1) > \mathcal{R}(\pi_2).
\end{equation}
This implies that in expectation both share the same optima:
\begin{equation}
 \argmax_{\pi} \mathds{E}\big[\hat{\mathcal{R}}(\pi \mid \mathcal{D})\big] = \argmax_{\pi} \mathcal{R}(\pi).
\end{equation}
Therefore, optimizing a ranking model $\pi$ w.r.t.\ $\hat{\mathcal{R}}$ is expected to optimize it w.r.t.\ $\mathcal{R}$ as well, thus allowing for unbiased optimization.
Previous work has introduced several methods for maximizing $\hat{\mathcal{R}}$ so as to optimize different \ac{LTR} metrics~\citep{joachims2017unbiased, oosterhuis2020topkrankings, agarwal2019counterfactual}.

\section{Method}
\label{sec:genspecltr}

This section introduces our \emph{\acl{GENSPEC}} (GEN\-SPEC) framework  for query-speciali\-zation in counterfactual \ac{LTR}.
First, we describe how a feature-based model is optimized for generalization across queries.
Second, we explain how we optimize tabular models for specialized performance on individual queries.
Third, we introduce novel high-confidence bounds on relative performance, allowing us to reliably estimate performance differences between models.
Fourth, we propose the \ac{GENSPEC} meta-policy that chooses between the optimized models using the novel high-confidence bounds.
Finally, the section concludes with a summary of the \ac{GENSPEC} framework.

\subsection{Feature-Based Query-Generalization}

Feature-based \ac{LTR} optimizes models that try to predict the optimal model from query-document features.
There are many feature-based models one can choose; in \ac{LTR} linear models, support-vector-machines, neural networks and decision tree ensembles are popular choices.
All of these models can be conceptualized as a function $f_\theta(q, d) \in \mathds{R}$ that transforms the features available for query $q$ and document $d$ into a score based on the learned parameters $\theta$.
Rankings are then constructed by sorting all documents $d$ for a query $q$ according to their predicted scores.

In practice, a feature-based ranking model often does not sort every document in the collection~\citep[see, e.g.,][]{wang-2011-cascade}; instead, earlier steps pre-select only a subset of documents that make it into the final ranking.
We use $(d, d') \in q$ as shorthand to denote that both documents $d$ and $d'$ have made it through the pre-selection.
Furthermore, we use $d \prec_y d'$ to indicate that $d$ is ranked higher than $d$ in ranking $y$.
To deal with ties between documents in their predicted scores, we introduce the set of valid rankings according to $f_\theta$:
\begin{align}
Y_\theta(q)  =  \Big \{ y  \mid  
  \forall (d, d') \in q, 
 \big(f_\theta(q, d) > f_\theta(q, d')
 \rightarrow d \prec_y d' \big)  \Big \}.
\end{align}
The final feature-based ranking model simply chooses uniform randomly from the set of valid rankings:
\begin{align}
\pi_\theta(y \mid q) =
\begin{cases}
\frac{1}{|Y_\theta(q)|} & \text{if } y \in Y_\theta(q),
\\
0 & \text{otherwise.}
\end{cases}
\end{align}
In theory, the optimal set of parameters maximizes the true reward:
\begin{align}
\theta^* = \argmax_{\theta} R(\pi_\theta).
\end{align}
In practice, counterfactual \ac{LTR} optimizes $\theta$ w.r.t.\ the estimated reward: $\hat{R}(\pi_\theta \mid \mathcal{D})$.
Many \ac{LTR} methods can be applied, we apply stochastic gradient descent on the counterfactual version of LambdaLoss~\citep{wang2018lambdaloss} as described by~\citet{oosterhuis2020topkrankings}.
Importantly, to avoid overfitting $\mathcal{D}$ is divided into a training and evaluation partition, a standard practice in machine learning.
The parameters $\theta$ are found using gradients based on the training partition, but chosen to maximize the expected reward on the validation partition.
As a result, $\theta$ is expected to maximize the reward on previously unseen queries.
This choice leads to a ranking model $\pi_\theta$ that has robust behavior that generalizes across queries.
Feature-based models are particularly suited for query-generalization because they can apply behavior learned on one query to any other query, including those previously unseen.

\subsection{Tabular Query-Specialization}

While feature-based models are good at generalization, they are often limited by the available features.
In many cases, the available features do not provide enough information to predict the optimal ranking.
This limitation does not apply to tabular models, which try to memorize the optimal rankings instead of predicting them from features~\citep{zoghi2016click}.
Because they do not depend on features, tabular models can cover all possible permutations of documents.
However, memorization does not generalize, i.e., the ranking behavior learned on one query cannot be applied to another.
Instead, tabular models are well suited for query-specialization: learning extremely high-performing behavior for individual queries.

As discussed in Section~\ref{sec:relatedwork}, many online \ac{LTR} algorithms optimize tabular models for the rankings of individual queries.
We will counterfactually estimate the relevance of every document based on the available data $\mathcal{D}$:
\begin{equation}
\hat{r}(q, d \mid \mathcal{D}) = \frac{1}{\sum_{i \in \mathcal{D}} \mathds{1}[q_i = q]} \sum_{i \in \mathcal{D}} \mathds{1}[q_i = q] \cdot \frac{c_i(d)}{\rho_i(d)},
\end{equation}
resulting in a tabular model that stores $\hat{r}(q, d \mid \mathcal{D})$ for every query-document pair observed in $\mathcal{D}$.
We note that this counterfactual estimate is also used by the \ac{PBM}~\citep{lagree2016multiple}, although  \ac{PBM} applies it to data gathered by its bandit algorithm.
Again, we use a set of valid rankings to deal with cases where multiple documents have the same estimated relevance:
\begin{equation*}
Y_\mathcal{D}(q)  = \Big \{ y  \,|\,  
  \forall (d, d') \in q, 
 \big(\hat{r}(q, d \,|\, \mathcal{D}) > \hat{r}(q, d' |\, \mathcal{D})
 \rightarrow d \prec_y d' \big)  \Big \}.
\end{equation*}
The resulting ranking model simply chooses uniform randomly from the set of valid rankings:
\begin{align}
\pi_\mathcal{D}(y \mid q) =
\begin{cases}
\frac{1}{|Y_\mathcal{D}(q)|} & \text{if } y \in Y_\mathcal{D}(q),
\\
0 & \text{otherwise.}
\end{cases}
\end{align}
Importantly, $\pi_\mathcal{D}$ always maximizes the estimated reward:
\begin{align}
\hat{R}(\pi_\mathcal{D} \mid \mathcal{D}) = \max_{\pi} \hat{R}(\pi \mid \mathcal{D}).
\label{eq:overfitreward}
\end{align}
This is both a benefit and a risk:
on the one hand, it means that $\pi_\mathcal{D}$ is optimal if the estimate is correct, i.e., $\mathcal{D}$ is large enough for an accurate $\hat{R}(\pi_\mathcal{D} \mid \mathcal{D}) $;
on the other hand, $\pi_\mathcal{D}$ is extremely sensitive to noise since it is completely overfitting on $\mathcal{D}$.
Therefore, if $\mathcal{D}$ is small and the estimate is probably highly incorrect due to noise, $\pi_\mathcal{D}$ is likely to have very poor performance.
For instance, in the extreme case that only a single click is available in $\mathcal{D}$ for a query, $\pi_\mathcal{D}$ will place the clicked document on top of the ranking for that query, despite a single click being very poor evidence of a document being relevant.
Moreover, if no clicks are available for a query, then $\pi_\mathcal{D}$ will uniform randomly select any possible permutation.

In stark contrast with the feature-based $\pi_\theta$, the clicks related to one query will never affect the behavior of $\pi_\mathcal{D}$ w.r.t.\ any other query.
Thus, one could also think of $\pi_\mathcal{D}$ as an ensemble of models each specialized in a single query.
This is very similar to the behavior of online \ac{LTR} algorithms that optimize tabular models, as they often see each query as an individual ranking problem.
Similarly, $\pi_\mathcal{D}$ can have extremely high performance for one query, while also displaying detrimental behavior for another.

\subsection{Reliably Choosing Between Models}

So far, we have introduced the feature-based ranking model $\pi_\theta$ and the tabular ranking model $\pi_\mathcal{D}$.
It also appears that there is no clear optimal choice between $\pi_\theta$ and $\pi_\mathcal{D}$; instead, this choice seems to mostly depend on the available data $\mathcal{D}$.
We wish to deploy the model that leads to the highest performance, however, we also want to avoid a detrimental user experience due to choosing the wrong model.
Recent work by \citet{jagerman2020safety} introduced the \acf{SEA}, for choosing safely between a safe model and a risky learned model.
\ac{SEA} applies high confidence bounds~\citep{thomas2015high} to the performances of both models.
The risky learned model is only deployed if the lower bound on its performance is greater than the upper bound of the safe initial model.
In other words, if with high confidence \ac{SEA} can conclude that deploying the risky model does not lead to a decrease in performance.

We want to apply a similar strategy to choose between $\pi_\theta$ and $\pi_\mathcal{D}$, however, we note that \ac{SEA} uses two bounds to make a single decision.
This is not necessary and instead we introduce a novel bound on the relative performance difference between two models.
By using only a single bound, the high-confidence decision can be made with considerably less data.

Let $\pi_1$ and $\pi_2$ be any two models, and
let $\delta(\pi_1, \pi_2)$ indicate the true difference in performance between them:
\begin{equation}
\delta(\pi_1, \pi_2) = \mathcal{R}(\pi_1) - \mathcal{R}(\pi_2).
\end{equation}
Knowing $\delta(\pi_1, \pi_2)$ allows us to optimally choose which of the two models to deploy.
However, we can only estimate its value from historical data $\mathcal{D}$:
\begin{equation}
\hat{\delta}(\pi_1, \pi_2 \mid \mathcal{D}) = \hat{\mathcal{R}}(\pi_1 \mid \mathcal{D}) - \hat{\mathcal{R}}(\pi_2 \mid \mathcal{D}).
\end{equation}
For brevity, let $R_{i,d}$ indicate the inverse-propensity-scored difference for a single document $d$ at interaction $i$:
\begin{equation}
R_{i,d} =  
\frac{c_i(d)}{\rho_i(d)} \sum_{y \in \pi_1 \cup \pi_2} \big(\pi_1(y \mid q_i) - \pi_2(y \mid q_i)\big) 
 \cdot  \lambda\big(\textit{rank}(d \mid y)\big).
\end{equation}
Then, for computational efficiency we rewrite:
\begin{equation}
\hat{\delta}(\pi_1, \pi_2 \mid \mathcal{D}) = 
 \frac{1}{|\data|} \sum_{i \in \mathcal{D}} \sum_{d \in y_i} R_{i,d} = \frac{1}{|\data|  K} \sum_{i \in \mathcal{D}} \sum_{d \in y_i} K \cdot R_{i,d}.
\end{equation}
With the confidence parameter $\epsilon \in [0, 1]$,
setting $b$ to be the maximum possible absolute value for $R_{i,d}$, i.e., $b = \frac{\max \lambda(\cdot)}{\min \rho}$, and
\begin{align}
\nu =  \frac{ 2  |\mathcal{D}|  K  \ln\big(\frac{2}{1-\epsilon}\big)}{|\mathcal{D}| K-1}  \sum_{i \in \mathcal{D}} \sum_{d \in y_i} \big(K \cdot R_{i,d} - \hat{\delta}(\pi_1, \pi_2 \mid \mathcal{D}) \big)^2,
\end{align}
we follow~\citet{thomas2015high} to get the high-confidence bound:
\begin{align}
\textit{CB}
(\pi_1, \pi_2 \mid \mathcal{D})
= \frac{7 K b\ln\big(\frac{2}{1-\epsilon}\big)}{3(|\mathcal{D}|  K-1)} + \frac{1}{|\mathcal{D}|  K}  
 \cdot \sqrt{\nu}. 
\label{eq:CB}
\end{align}
In turn, this provides us with the following upper and lower confidence bounds on $\delta$:
\begin{equation}
\begin{split}
\textit{LCB}(\pi_1, \pi_2 \mid \mathcal{D}) &= \hat{\delta}(\pi_1, \pi_2 \mid \mathcal{D}) - \textit{CB}(\pi_1, \pi_2 \mid \mathcal{D})  \\
\textit{UCB}(\pi_1, \pi_2 \mid \mathcal{D}) &= \hat{\delta}(\pi_1, \pi_2 \mid \mathcal{D}) + \textit{CB}(\pi_1, \pi_2 \mid \mathcal{D}). 
\end{split}
\label{eq:ucb}
\end{equation}
As proven by~\citet{thomas2015high}, with at least a probability of $\epsilon$ they bound the true value of $\delta(\pi_1, \pi_2)$:
\begin{equation}
P\Big(\delta(\pi_1, \pi_2) \in \big[\textit{LCB}(\pi_1, \pi_2 \mid \mathcal{D}), \textit{UCB}(\pi_1, \pi_2 \mid \mathcal{D})\big]
 \Big) > \epsilon. \label{eq:safetygaurantee}
\end{equation}
In other words, if $\textit{LCB}(\pi_1, \pi_2 \mid \mathcal{D}) > 0$, the probability that $\pi_1$ outperforms $\pi_2$ is higher than $\epsilon$, i.e., $P(\mathcal{R}(\pi_1) > \mathcal{R}(\pi_2)) > \epsilon$.

We note that the novelty of our bound is that it bounds the \emph{performance difference of two models}, instead of the \emph{performance of individual models} used in earlier work by \citet{jagerman2020safety} and \citet{thomas2015high}.
Whereas the \ac{SEA} approach uses two bounds, our approach can decide between models with only a single bound.
In Appendix~\ref{sec:theoryrelativebounds} we theoretically analyze the difference between these approaches and conclude that our relative bound is more data-efficient if there is a positive covariance between $ \hat{\mathcal{R}}(\pi_1 \mid \mathcal{D})$ and $ \hat{\mathcal{R}}(\pi_2 \mid \mathcal{D})$.
Because both estimates are based on the same interaction data $\mathcal{D}$, a high covariance is extremely likely.
In practice, this means that our novel bound requires much less data to identify a better performing model than \ac{SEA}.

\begin{figure}[t]
\centering
\includegraphics[clip,trim=0mm 1mm 0mm 0mm,width=0.75\columnwidth]{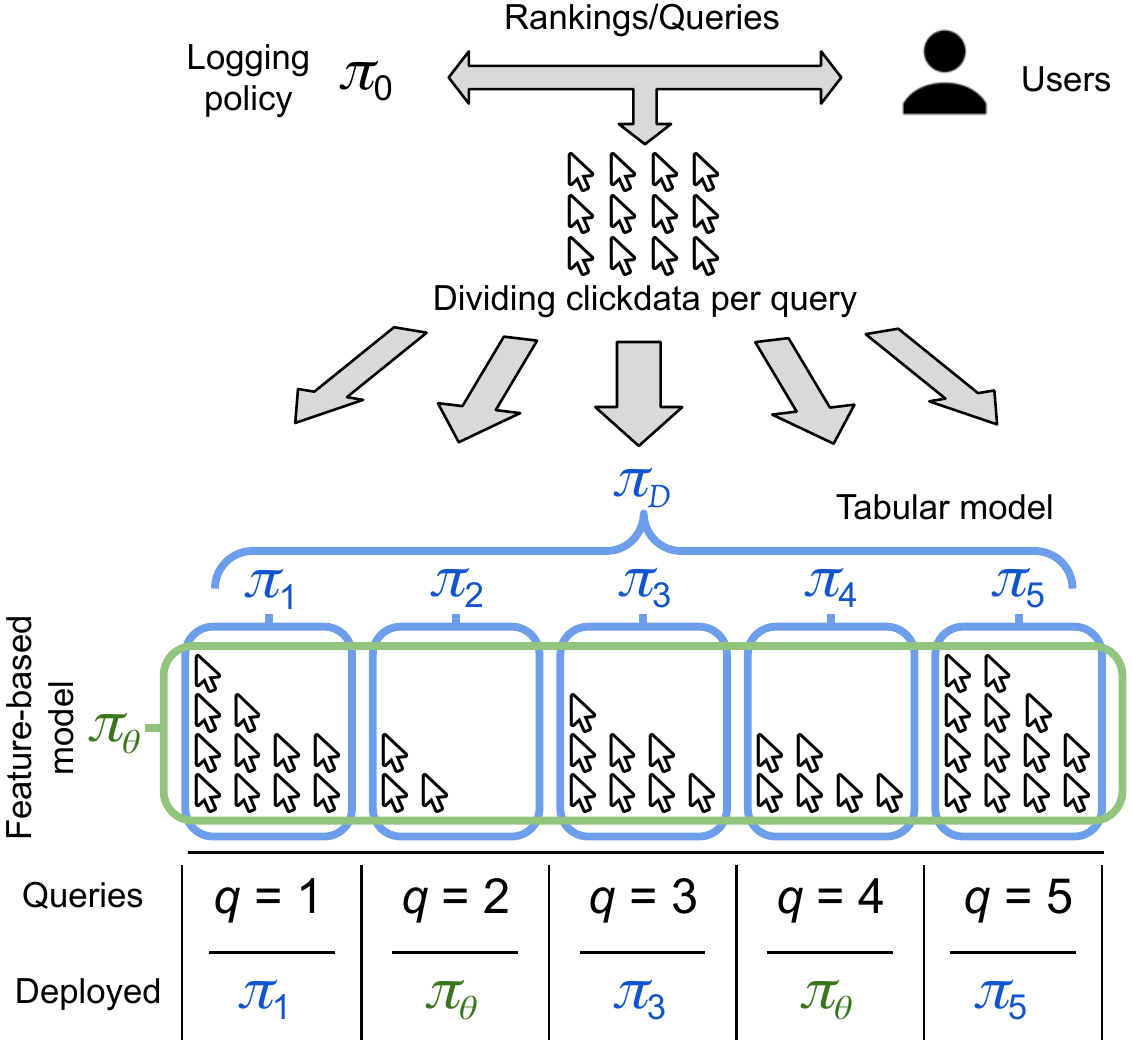} \\
\caption{
Visualization of the \ac{GENSPEC} framework. %
A feature-based model $\pi_\theta$ is trained on the complete dataset,
the tabular model $\pi_\mathcal{D}$ consists of
many specialized models $\pi_1, \pi_2, \ldots$ each highly-specialized for a single query.
\ac{GENSPEC} decides which model to deploy per query, based on high-confidence bounds.
}
\vspace{-0.9\baselineskip}
\label{fig:genspec}
\end{figure}

\subsection{\acl{GENSPEC}}

So far we have described the four main ingredients of the \ac{GENSPEC} Framework:
\begin{enumerate*}[label=(\roman*)]
\item the logging policy $\pi_0$ used to gather the click data $\mathcal{D}$, $\pi_0$ is assumed to be safe to deploy w.r.t.\ user experience;
\item the feature-based ranking model $\pi_\theta$ optimized for robust performance generalized across queries;
\item the tabular ranking model $\pi_\mathcal{D}$ that specializes by memorizing a ranking per query; and
\item the novel high-confidence bound on model performance differences that can be used to choose between models.
\end{enumerate*}
\ac{GENSPEC} will choose between deploying $\pi_0$, $\pi_\theta$, and $\pi_\mathcal{D}$ on a per query basis by applying a doubly conservative strategy:
$\pi_\theta$ is only deployed when there is high-confidence that it outperforms $\pi_0$ across all queries;
$\pi_\mathcal{D}$ is only deployed for a query $q$ when there is high-confidence that it outperforms $\pi_0$ and $\pi_\theta$ on the same query $q$.

However, to avoid overfitting we should not use the same data $\mathcal{D}$ for both the training of models and to compute performance bounds.
This is especially important because $\pi_\mathcal{D}$ will completely overfit on $\mathcal{D}$, thus if evaluated on the same $\mathcal{D}$ the performance of $\pi_\mathcal{D}$ will always appear optimal (Eq.~\ref{eq:overfitreward}).
To avoid this overfitting problem, we split $\data$ in a training partition $\traindata$ and a model-selection partition $\bounddata$ so that 
$
\data = \traindata \cup \bounddata
$
and
$
\traindata \cap \bounddata = \emptyset
$.
For each model, we train two versions: one trained on the entire dataset $\mathcal{D}$, and another only trained on $\traindata$.
Let $\pi_\theta'$ and $\pi_\mathcal{D}'$ indicate the versions of feature-based model and the tabular model trained on $\traindata$.
These models can now safely be compared on the $\bounddata$ partition without risk of overfitting.

The \ac{GENSPEC} strategy assumes that the performance of $\pi_\theta$ is always greater than $\pi_\theta'$: 
$\mathcal{R}(\pi_\theta) > \mathcal{R}(\pi_\theta')$, and similarly: $\mathcal{R}(\pi_\mathcal{D}) > \mathcal{R}(\pi_\mathcal{D}')$.
This is a reasonable assumption since $\pi_\theta$ is trained on a superset of the data on which $\pi_\theta'$ is trained.
Using this assumption, \ac{GENSPEC} chooses between the feature-based model $\pi_\theta$ and the logging policy $\pi_0$, using bounds computed on $\pi_\theta'$:
\begin{equation}
\pi_{G}(y \mid q)
= 
\begin{cases}
\pi_\theta(y \mid q),& \text{if } \textit{LCB}(\pi_\theta', \pi_0 \mid  \bounddata) > 0, \\
\pi_0(y \mid q),& \text{otherwise}. \\
\end{cases}
\label{eq:genspecchoice1}
\end{equation}
Therefore, $\pi_{G} = \pi_\theta$ only if with high confidence $\pi_\theta'$ outperforms $\pi_0$, otherwise $\pi_{G} =\pi_0$, i.e., the safe option is chosen.
By using $\pi_\theta'$ to decide whether to choose $\pi_\theta$ over $\pi_0$, we avoid the overfitting problem while also utilizing the expected higher performance of $\pi_\theta$ over $\pi_\theta'$.

The next step is to decide between deploying $\pi_\mathcal{D}$ and $\pi_{G}$.
Again to avoid overfitting, we use $\pi_{G}'$ for a copy of $\pi_{G}$ that is only based on $\traindata$:
\begin{equation}
\pi_{G}'(y \mid q)
= 
\begin{cases}
\pi_\theta'(y \mid q),& \text{if } \textit{LCB}(\pi_\theta', \pi_0 \mid  \bounddata) > 0, \\
\pi_0(y \mid q),& \text{otherwise}. \\
\end{cases}
\end{equation}
Remember that the behavior of tabular ranking models like $\pi_\mathcal{D}$ is independent per query, i.e., clicks related to one query will never affect the ranking behavior of $\pi_\mathcal{D}$ w.r.t.\ any other query.
For this reason, the choice between $\pi_\mathcal{D}$ and $\pi_{G}$ is made on a per query basis, unlike the choice between the generalized $\pi_\theta$ and safe $\pi_0$.
To do so, we divide $\bounddata$ per query, resulting in a $\bounddata_q$ per query $q$:
\begin{equation}
\bounddata_q = \big\{(c_i, y_i, \rho_i, q_i) \in \bounddata \mid q_i = q \big\}.
\end{equation}
This allows us to bound the relative performance between $\pi_\mathcal{D}'$ and  $\pi_{G}'$ w.r.t. a single query $q$.
Finally, we can use this to choose between deploying the highly-specialized $\pi_\mathcal{D}$ and the robust generalized $\pi_{G}$ per query:
\begin{equation}
\pi_{GS}(y \mid q)
= 
\begin{cases}
\pi_\mathcal{D}(y \mid q),& \text{if } \textit{LCB}(\pi_\mathcal{D}', \pi_{G}' \mid  \bounddata_q) > 0, \\
\pi_{G}(y \mid q),& \text{otherwise}. \\
\end{cases}
\label{eq:genspecchoice2}
\end{equation}
Thus $\pi_{GS}(y \mid q) = \pi_\mathcal{D}(y \mid q)$ only if there is high-confidence that $\pi_\mathcal{D}'$ outperforms  $\pi_{G}$.
Because this decision is made independently per query, it is entirely possible that for one query $\pi_{GS}$ deploys  $\pi_\mathcal{D}$ while deploying $\pi_{G}$ for another.
This behavior allows $\pi_{GS}$ to have the high performance of a successfully specialized $\pi_\mathcal{D}$ on queries where it is highly confident of such high performance, while also relying on the robust behavior of $\pi_\theta$ where this confidence has not been obtained.
We note that because decisions are only made with confidence, there will be some delay between the moment that $\pi_\mathcal{D}$ outperforms $\pi_{G}$ and when $\pi_\mathcal{D}$ is deployed.
Nevertheless, \ac{GENSPEC} safely combines the strongest advantages of each model: the safe behavior of $\pi_0$, the robust generalization of feature-based $\pi_\theta$ and high-performance at convergence of the tabular model $\pi_\mathcal{D}$.

\begin{algorithm}
\caption{The \ac{GENSPEC} training and serving procedures.}\label{alg:genspec}
\begin{algorithmic}[1]
\Procedure{Initialize}{$\pi_0, \mathcal{D}, \epsilon, \beta$} %
\label{alg:line:init}
\State $\traindata, \bounddata \gets \text{random\_split}(\mathcal{D}, \beta)$ \label{alg:line:datasep}
\State $\pi_\theta' \gets \text{train\_feature\_based\_model}(\traindata)$ \label{alg:line:trainfeat1}
\State $\pi_\mathcal{D}' \gets \text{infer\_tabular\_model}(\traindata)$ \label{alg:line:traintab1}
\State $\text{feat\_model\_activated} \gets \text{LCB}(\pi_\theta', \pi_0 \mid \bounddata) > 0$ \label{alg:line:bound1}
\State $\text{override\_queries} \gets \{\}$ \label{alg:line:emptyset} %
\For{$q \in \mathcal{D}$}\Comment{\emph{Loop over unique queries in $\mathcal{D}$.}}
\If{$\text{feat\_model\_activated}$}
\If{$\text{LCB}(\pi_\mathcal{D}', \pi_\theta' \mid \bounddata_q) > 0$} \label{alg:line:bound2}
\State $\text{override\_queries} \gets \text{override\_queries} \cup \{q\}$  \label{alg:line:add1}
\EndIf
\ElsIf{$\text{LCB}(\pi_\mathcal{D}', \pi_0 \mid \bounddata_q) > 0$} \label{alg:line:bound3}
\State $\text{override\_queries} \gets \text{override\_queries} \cup \{q\}$ \label{alg:line:add2}
\EndIf
\EndFor
\State $\pi_\theta \gets \text{train\_feature\_based\_model}(\mathcal{D})$ \label{alg:line:trainfeat2}
\State $\pi_\mathcal{D} \gets \text{infer\_tabular\_model}(\mathcal{D})$ \label{alg:line:traintab2}
\EndProcedure
\Procedure{Model\_to\_Serve}{$q$} \Comment{\emph{User-issued query $q$.}} \label{alg:line:input2}
\If{$q \in \text{override\_queries}$}  \label{alg:line:override}
\State \textbf{return} $\pi_\mathcal{D}$
\ElsIf{$\text{feat\_model\_activated}$} \label{alg:line:activated}
\State \textbf{return} $\pi_\theta$
\Else \label{alg:line:safe}
\State \textbf{return} $\pi_0$
\EndIf
\EndProcedure
\end{algorithmic}
\end{algorithm}

\subsection{Summary}

This completes our introduction of the \ac{GENSPEC} framework,
Algorithm~\ref{alg:genspec} summarizes it in pseudocode and
 Figure~\ref{fig:genspec} provides a visualization.
The implementation in Algorithm~\ref{alg:genspec} is divided in two procedures: intialization and serving.
The initialization phase trains the models and decides where they will be deployed, subsequently, the serving procedure shows how a model is selected for any incoming user-issued query.

The initialization procedure takes as input: 
\begin{enumerate*}[label=(\roman*)]
\item the safe logging policy $\pi_0$, 
\item the clickdata $\mathcal{D}$,  
\item the confidence parameter $\epsilon$, and 
\item $\beta$ the percentage of data to be held-out for $\bounddata$ (Line~\ref{alg:line:init}).
\end{enumerate*}
First, $\mathcal{D}$ is divided into training data $\traindata$ and selection data $\bounddata$ (Line~\ref{alg:line:datasep}),
then the feature-based model $\pi_\theta'$ and the tabular model $\pi_\mathcal{D}'$ are trained on $\traindata$ (Line~\ref{alg:line:trainfeat1} and~\ref{alg:line:traintab1}).
The feature-based model is activated if the lower confidence bound on the performance difference between $\pi_\theta'$ and $\pi_0$  is positive (Line~\ref{alg:line:bound1}).
Next, an empty set is created to keep track of the queries for which the tabular model will override the other models (Line~\ref{alg:line:emptyset}).
For each unique query $q$ in $\mathcal{D}$, the lower confidence bound on the performance difference for $q$ between $\pi_\mathcal{D}'$ and the activated model is computed (Line~\ref{alg:line:bound2} or~\ref{alg:line:bound3}).
If this bound is positive, the query is added to the override set (Line~\ref{alg:line:add1} or~\ref{alg:line:add2}).
Finally, another feature-based model $\pi_\theta$ and tabular model $\pi_\mathcal{D}$ are trained, this time on all available data $\mathcal{D}$ (Line~\ref{alg:line:trainfeat2} and~\ref{alg:line:traintab2}).

The serving procedure takes as input an incoming user-issued query $q$.
If $q$ is in the override set (Line~\ref{alg:line:override}) then $\pi_\mathcal{D}$ is used to generate the ranking;
if $q$ is not in the override set but the feature-based model is activated then $\pi_\theta$ is used (Line~\ref{alg:line:activated});
otherwise the safe option $\pi_0$ is used (Line~\ref{alg:line:safe}).

By choosing between the deployment of two types of models: one for generalization and another for specialization,
 \ac{GENSPEC} safely combines the high performance at convergence of specialization and the robust safe performance of generalization.
In this paper, we discuss how \ac{GENSPEC} can be used for query-specialization, however, other choices for models and the specialization task can be made.
For example, one could choose to optimize many models specialized in the preferences of individual users, and a single feature-based model for generalization across all users.
By choosing between models on a per user basis, \ac{GENSPEC} could be used for safe personalization.
Due to this flexibility, we refer to \ac{GENSPEC} as a \emph{framework} that can be used for a wide variety of safe specialization scenarios.

\setlength{\tabcolsep}{0.15em}

{\renewcommand{\arraystretch}{0.63}
\begin{figure*}[t]
\centering
\begin{tabular}{c r r r r}
&
 \multicolumn{1}{c}{  Yahoo! Webscope}
&
 \multicolumn{1}{c}{  MSLR-WEB30k}
&
 \multicolumn{1}{c}{  Istella}
 &
\multirow{10}{*}[-10.7em]{
\includegraphics[scale=.35]{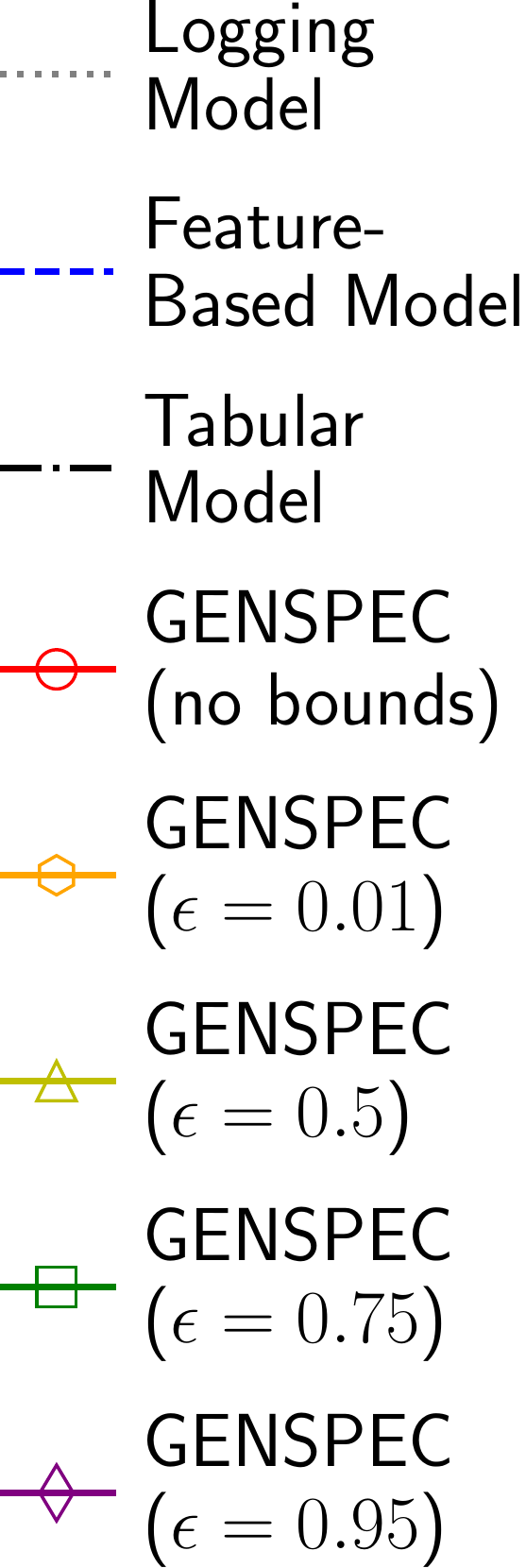}
} 
\\
\cmidrule{2-4}
& \multicolumn{3}{c}{Clicks generated with $\alpha=0.2$.}
\\
\cmidrule{2-4}
\rotatebox[origin=lt]{90}{\hspace{0.5em} \small Train-NDCG} &
\includegraphics[scale=0.38]{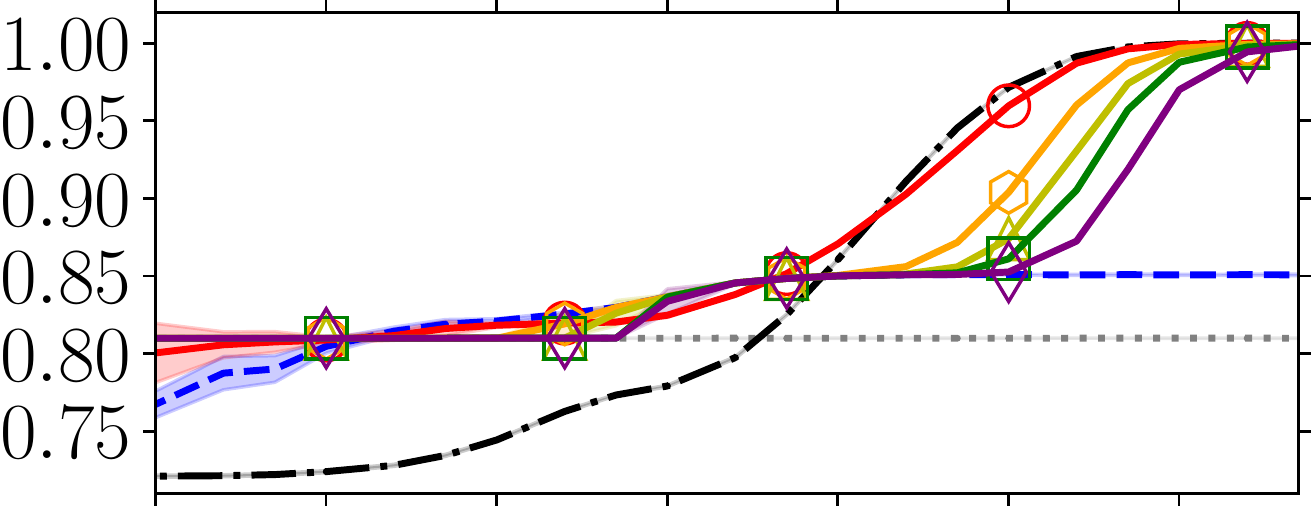} &
\includegraphics[scale=0.38]{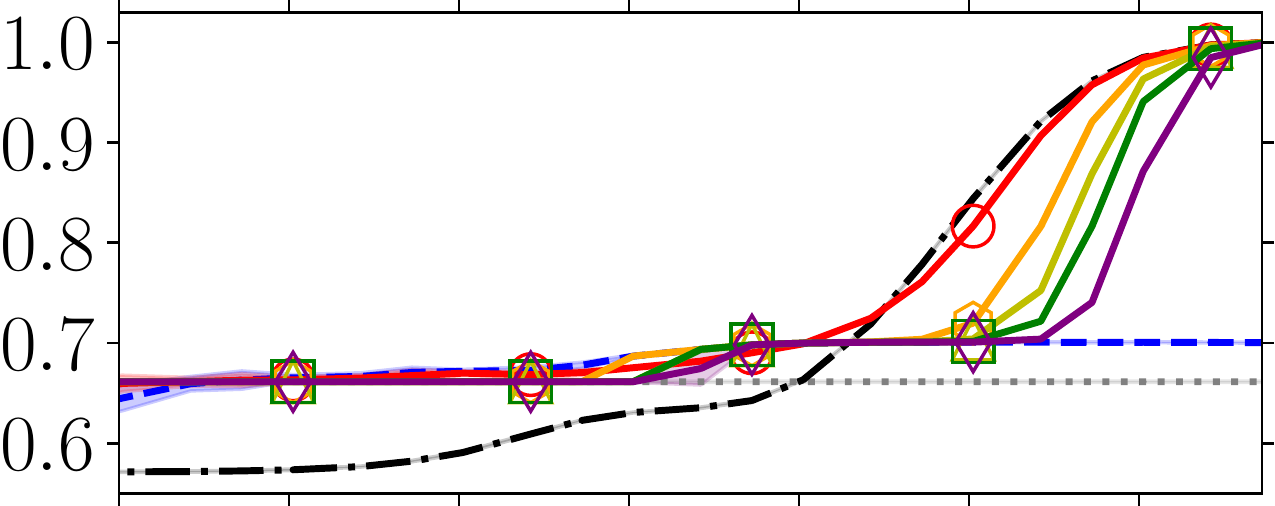} &
\includegraphics[scale=0.38]{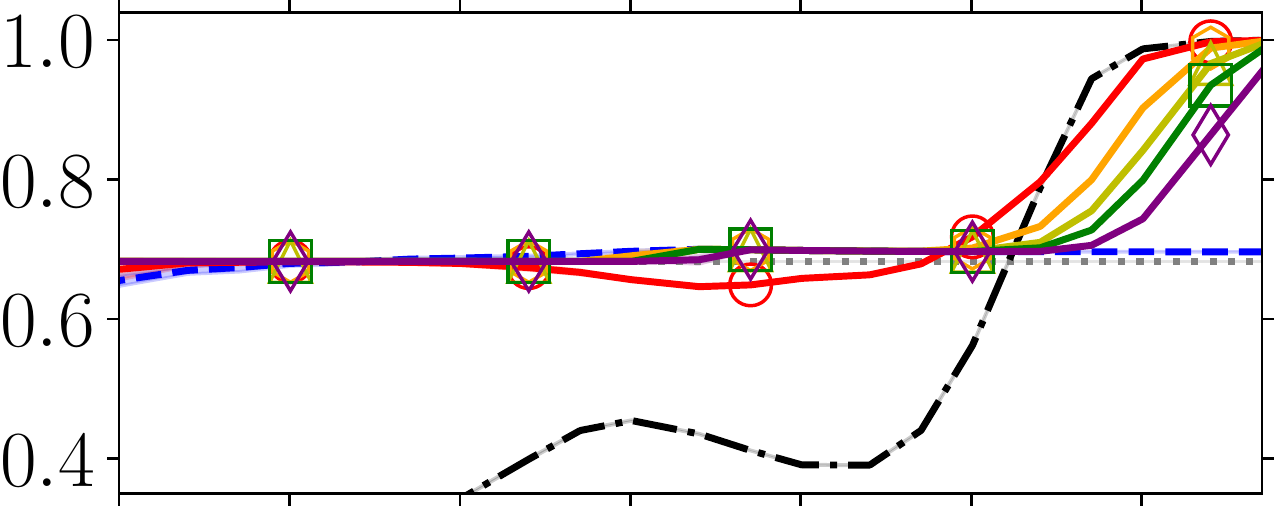} 
\\
\rotatebox[origin=lt]{90}{\hspace{1.5em} \small  Test-NDCG} &
\includegraphics[scale=0.38]{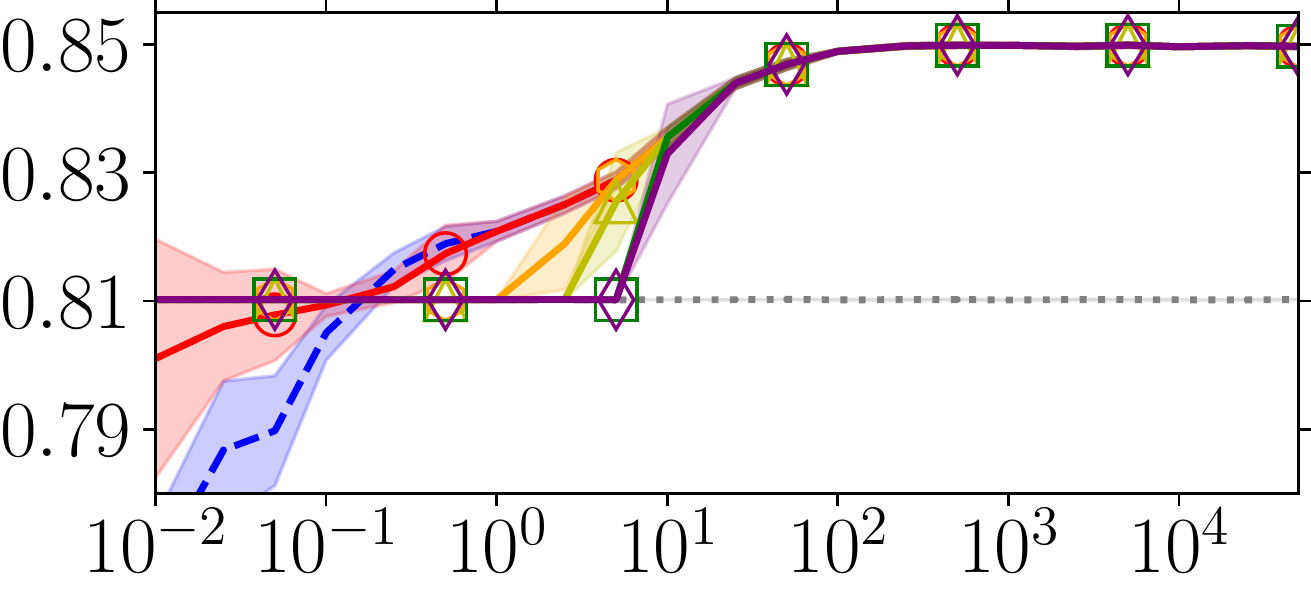} &
\includegraphics[scale=0.38]{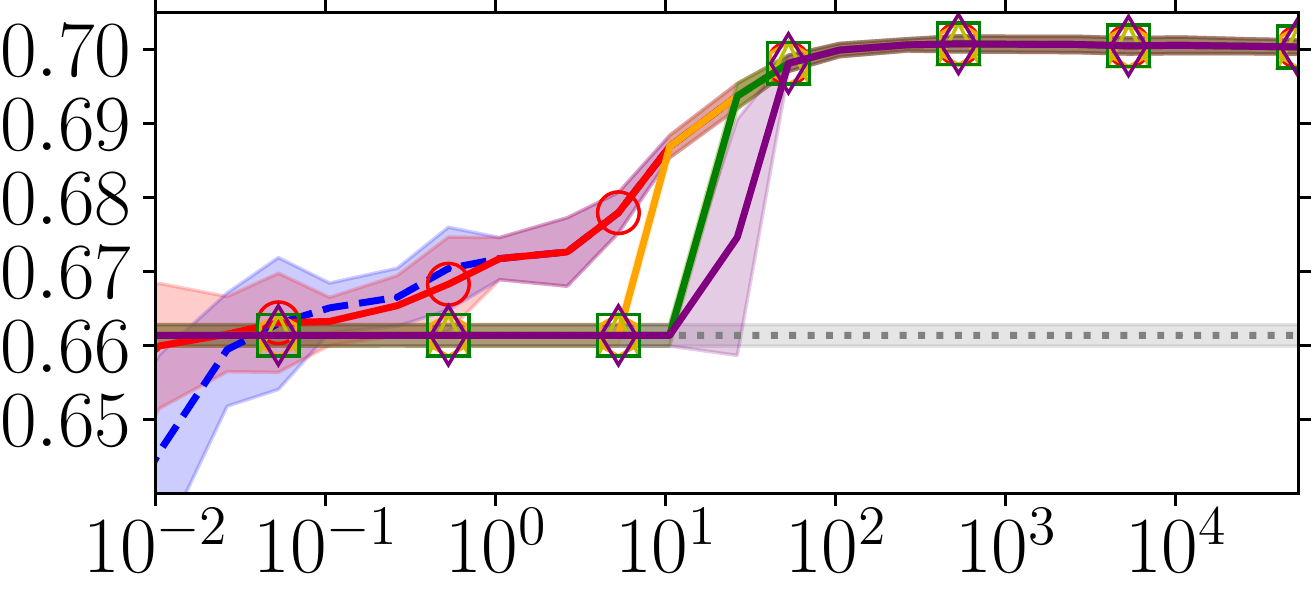} &
\includegraphics[scale=0.38]{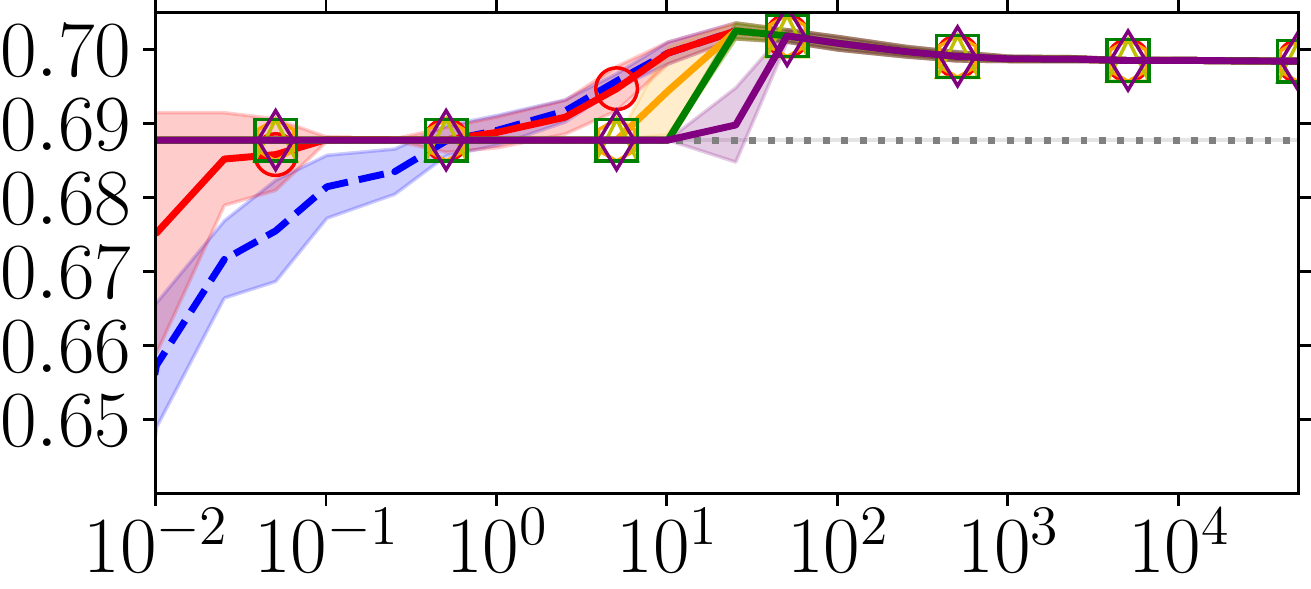}
\\
\cmidrule{2-4}
& \multicolumn{3}{c}{Clicks generated with $\alpha=0.025$.}
\\
\cmidrule{2-4}
\rotatebox[origin=lt]{90}{\hspace{0.5em} \small Train-NDCG} &
\includegraphics[scale=0.38]{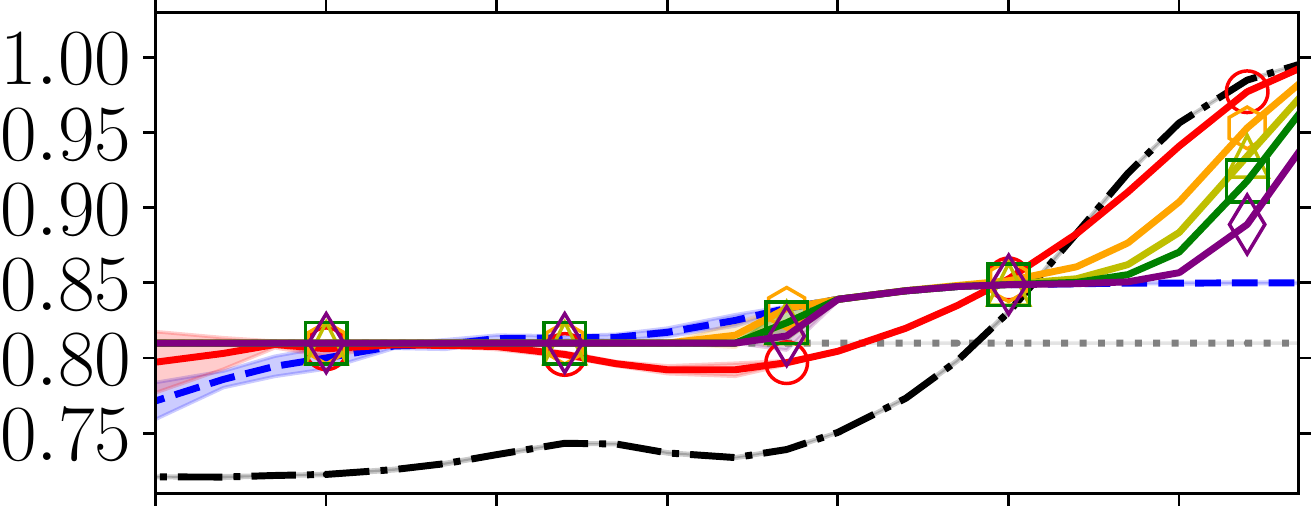} &
\includegraphics[scale=0.38]{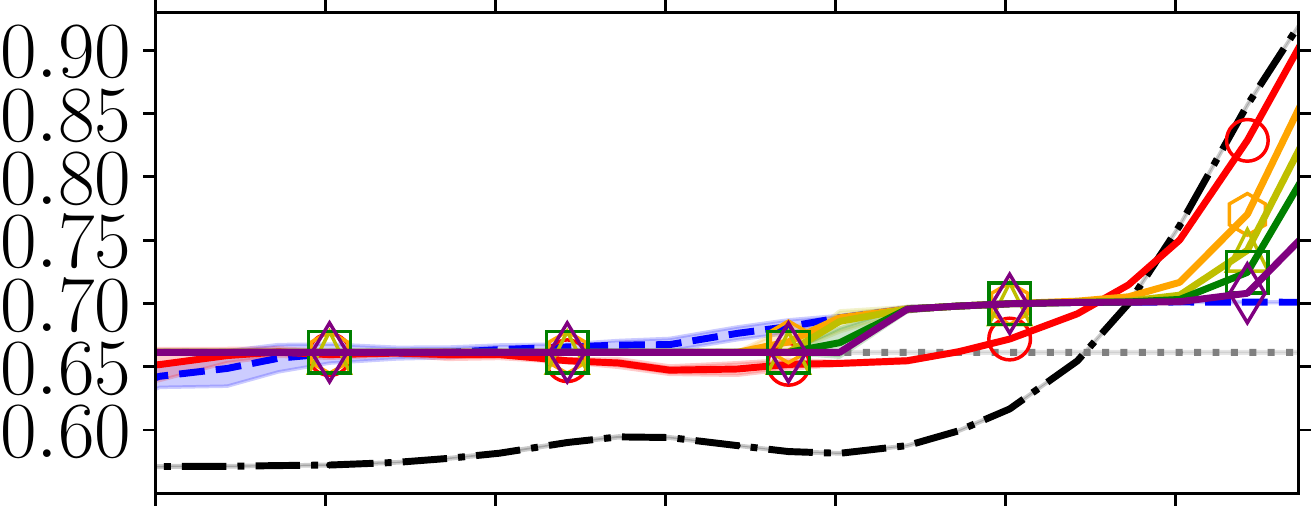} &
\includegraphics[scale=0.38]{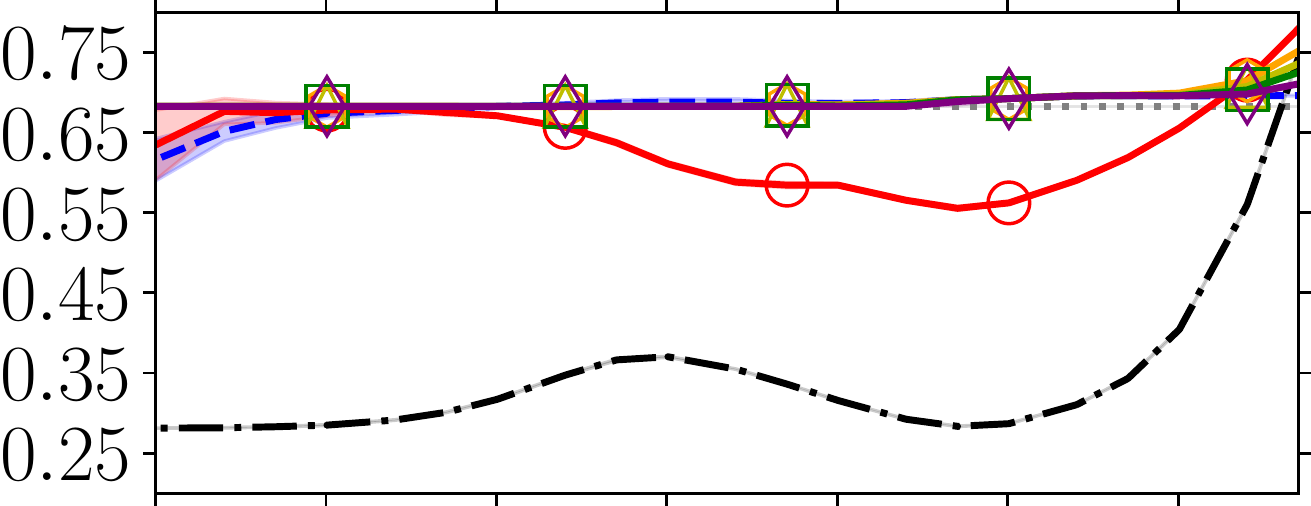} 
\\
\rotatebox[origin=lt]{90}{\hspace{1.5em} \small  Test-NDCG} &
\includegraphics[scale=0.38]{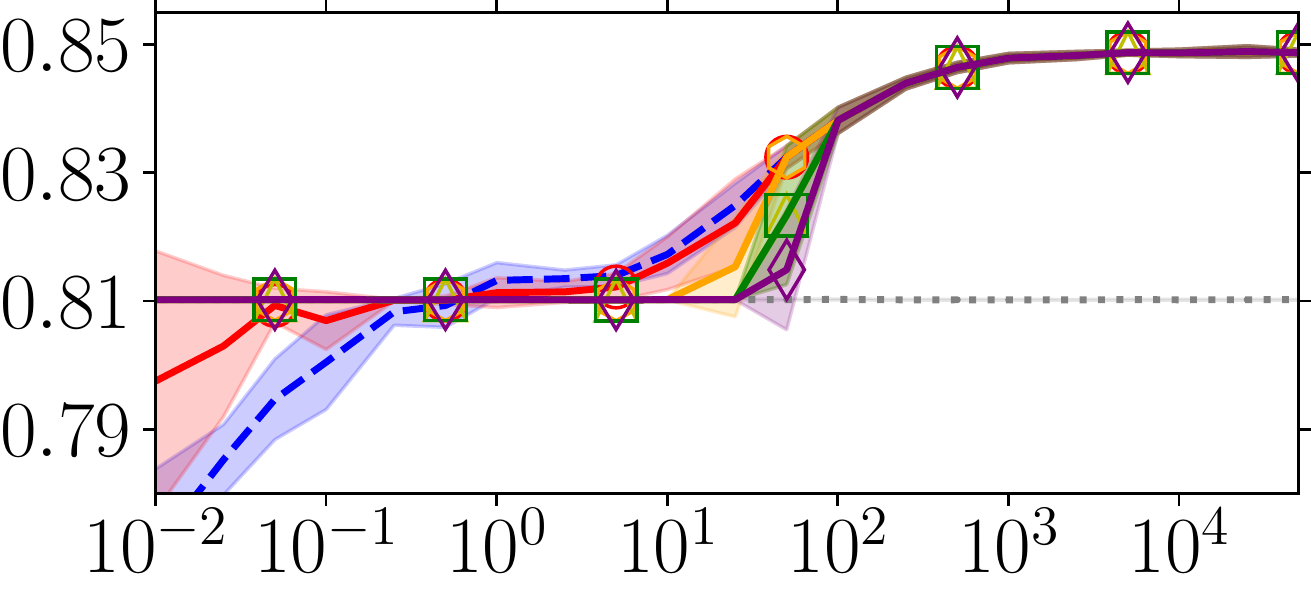} &
\includegraphics[scale=0.38]{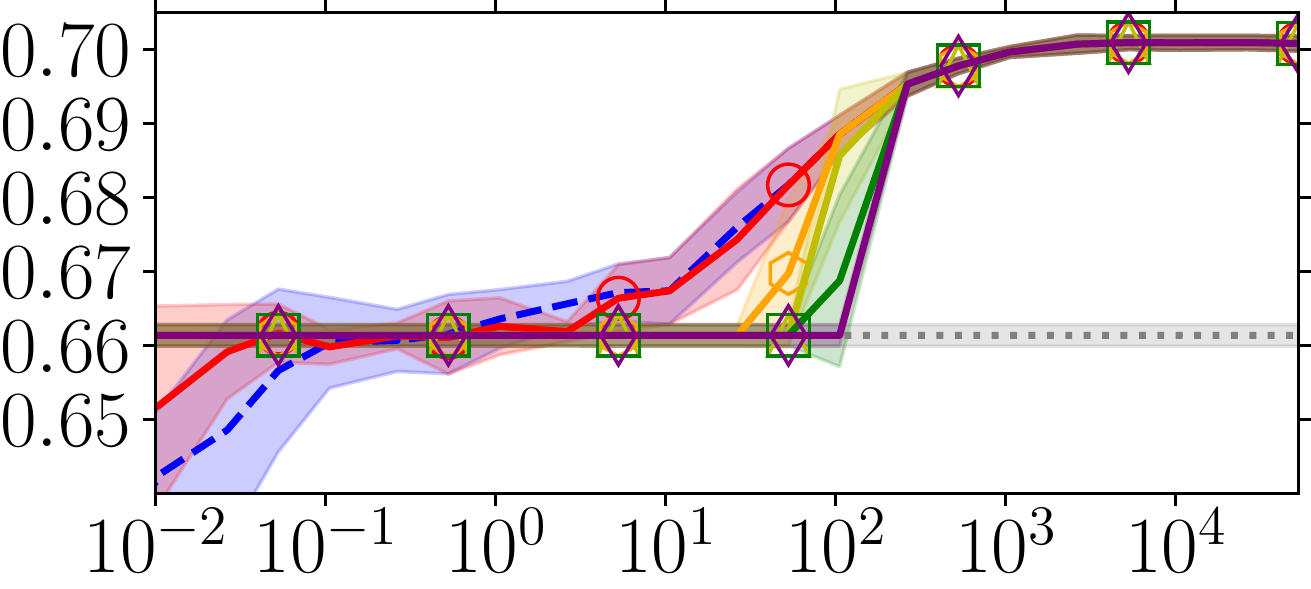} &
\includegraphics[scale=0.38]{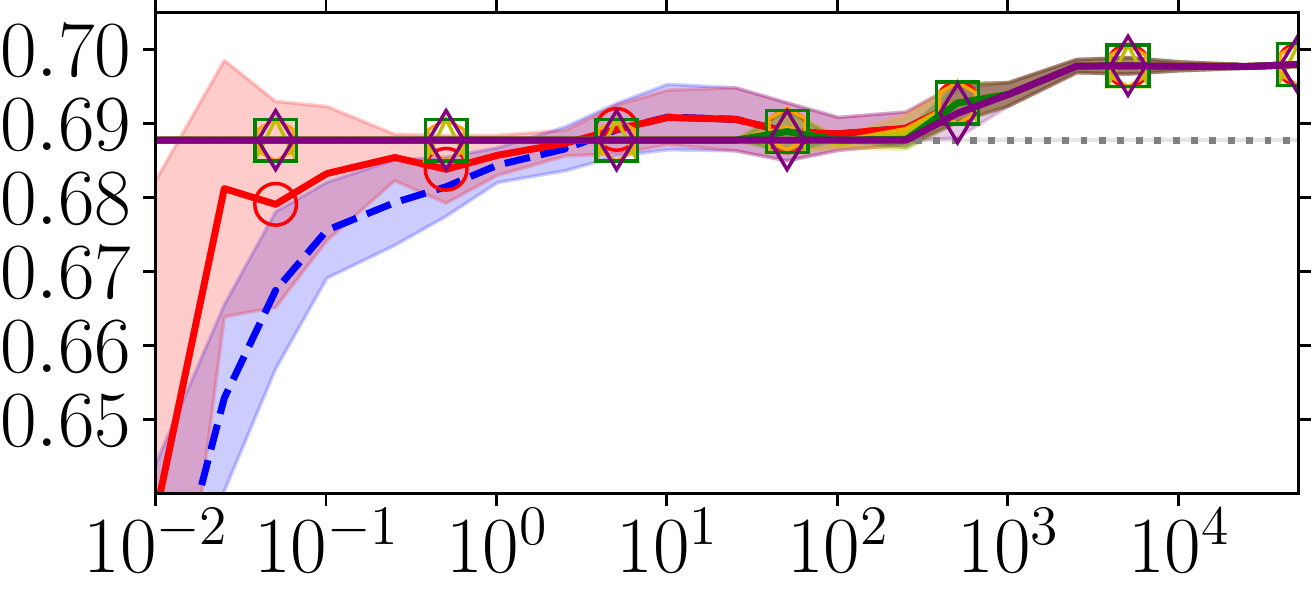}
\\
& \multicolumn{1}{c}{\small \hspace{0.5em} Mean Number of Clicks per Query}
& \multicolumn{1}{c}{\small \hspace{0.5em} Mean Number of Clicks per Query}
& \multicolumn{1}{c}{\small \hspace{0.5em} Mean Number of Clicks per Query}
\end{tabular}
\vspace{0.3\baselineskip}
\caption{
Performance of \ac{GENSPEC} with varying levels of confidence, compared to pure generalization and pure specialization.
We separate queries on the training set (Train-NDCG) that have received clicks, and queries on the test set (Test-NDCG) that do not receive any clicks.
Clicks are spread uniformly over the training set, the x-axis indicates the total number of clicks divided by the number of training queries.
Results are an average of 10 runs; shaded area indicates the standard deviation.
}
\label{fig:mainresults}
\end{figure*}
}

{\renewcommand{\arraystretch}{0.63}
\begin{figure*}[t]
\centering
\begin{tabular}{c r r r r}
&
 \multicolumn{1}{c}{  Yahoo! Webscope}
&
 \multicolumn{1}{c}{  MSLR-WEB30k}
&
 \multicolumn{1}{c}{  Istella}
 &
\multirow{10}{*}[-12em]{
\includegraphics[scale=.35]{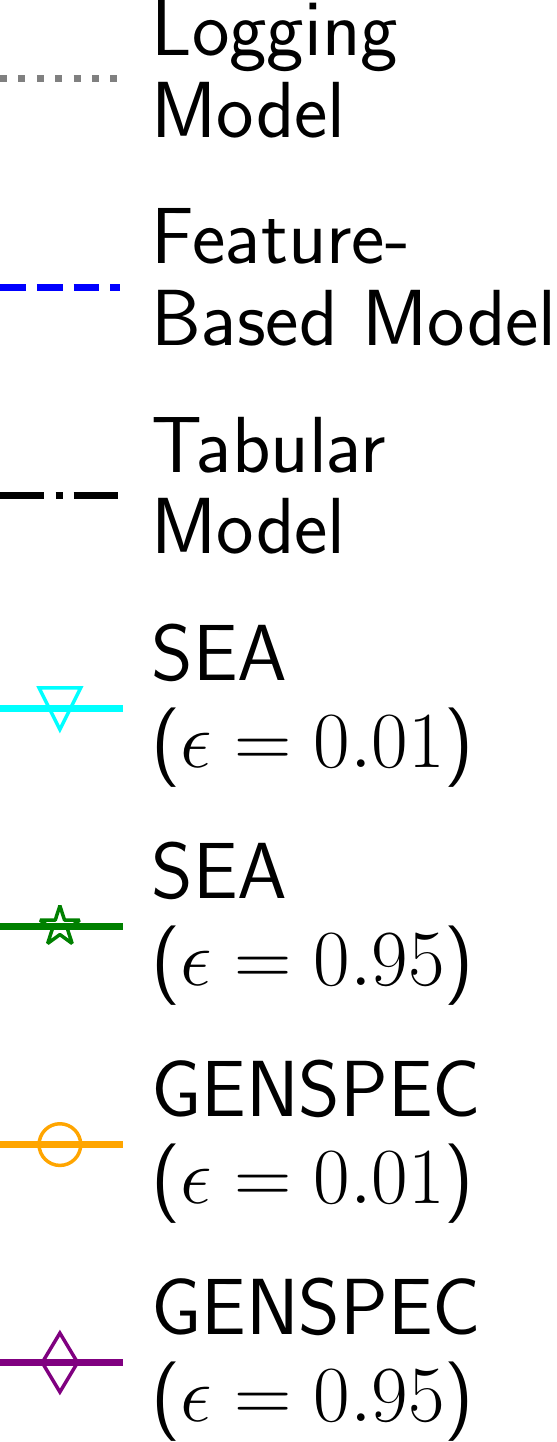}
} 
\\
\cmidrule{2-4}
& \multicolumn{3}{c}{Clicks generated with $\alpha=0.2$.}
\\
\cmidrule{2-4}
\rotatebox[origin=lt]{90}{\hspace{0.5em} \small Train-NDCG} &
\includegraphics[scale=0.38]{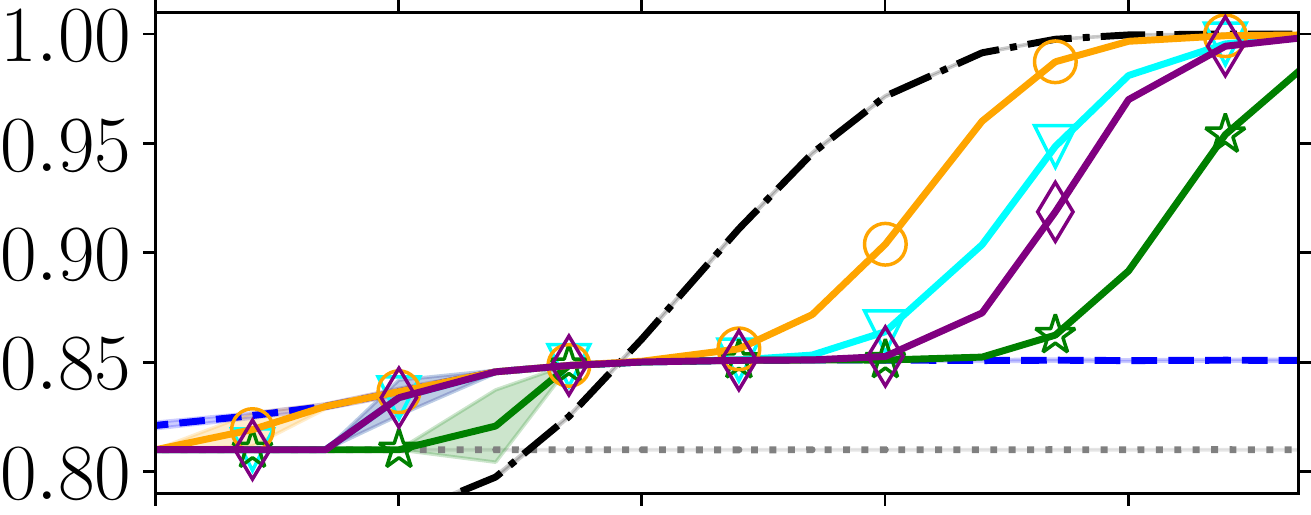} &
\includegraphics[scale=0.38]{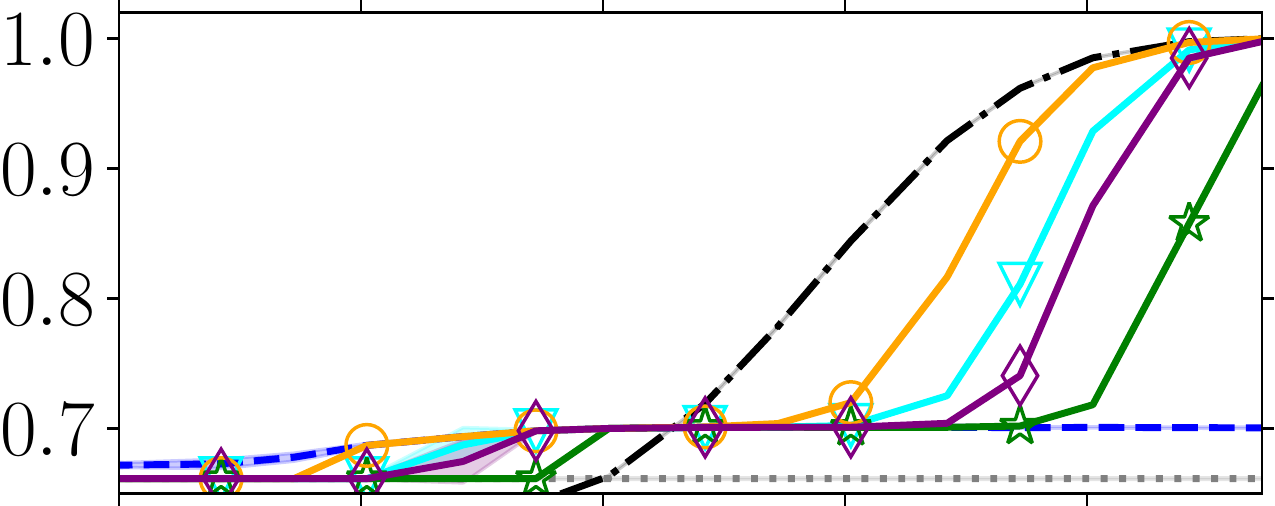} &
\includegraphics[scale=0.38]{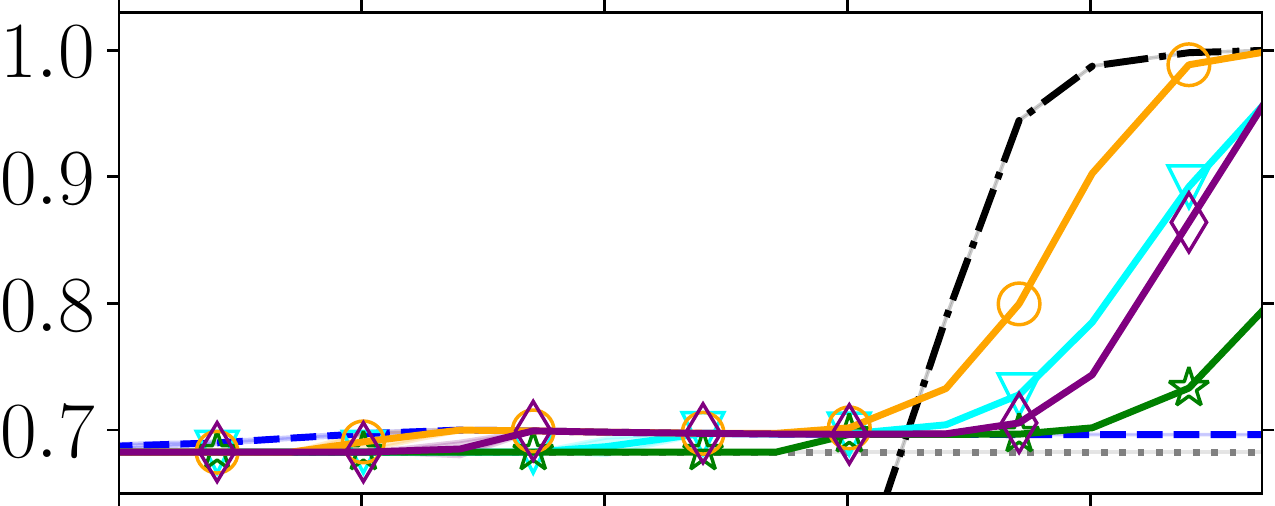} 
\\
\rotatebox[origin=lt]{90}{\hspace{1.5em} \small  Test-NDCG} &
\includegraphics[scale=0.38]{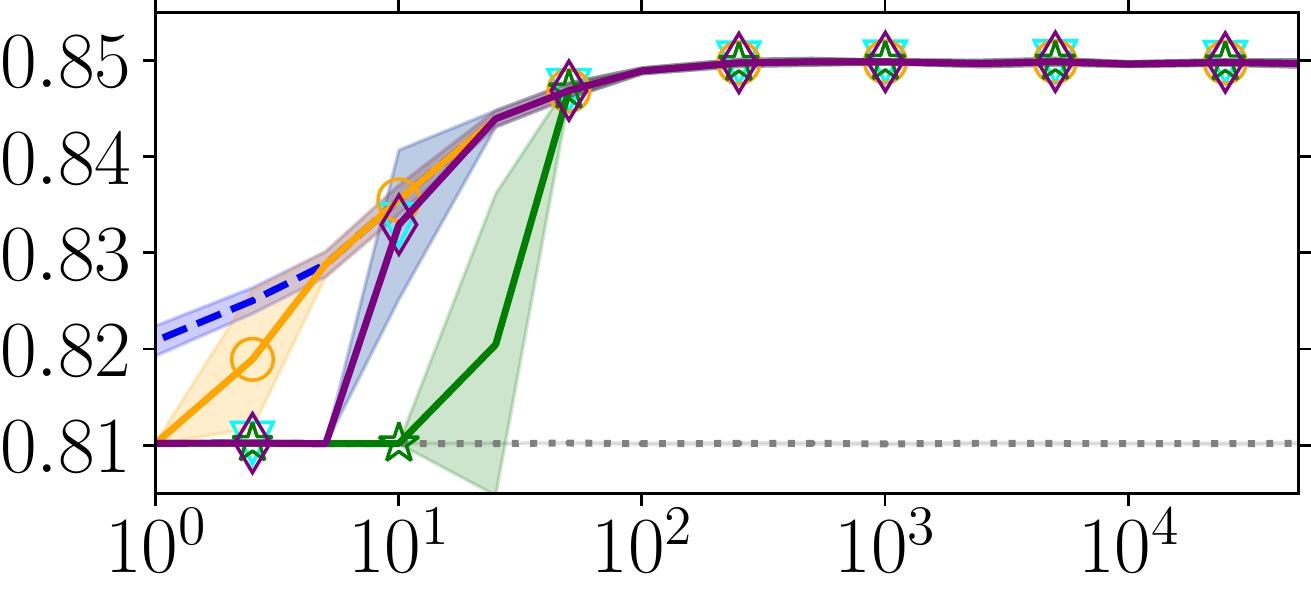} &
\includegraphics[scale=0.38]{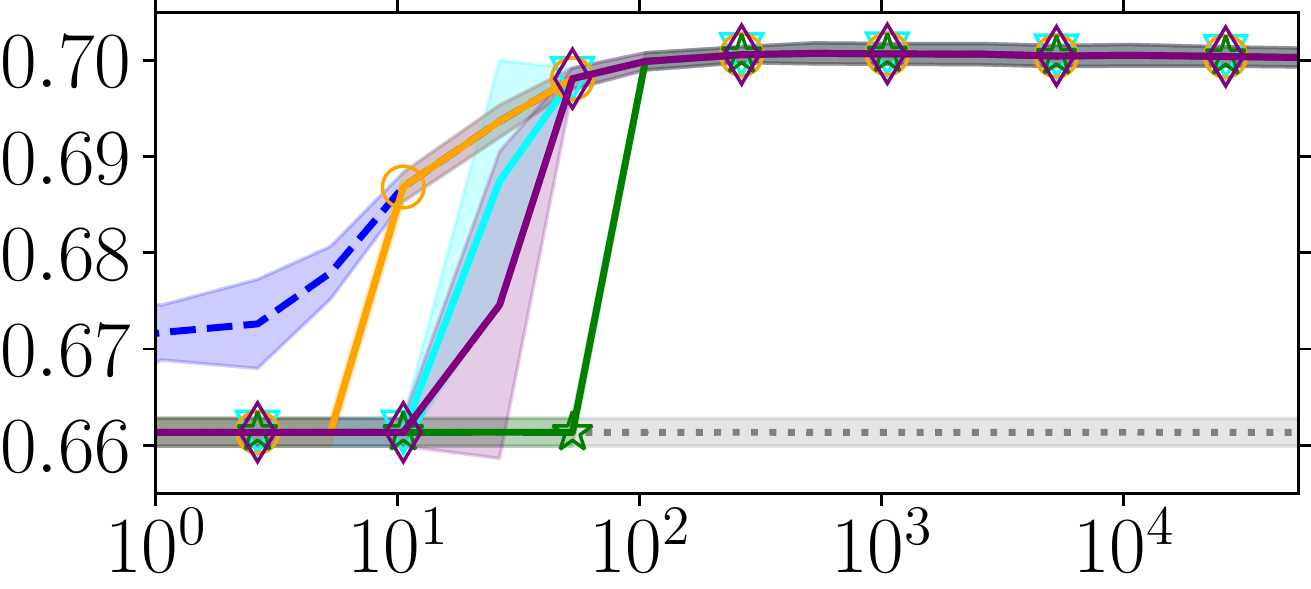} &
\includegraphics[scale=0.38]{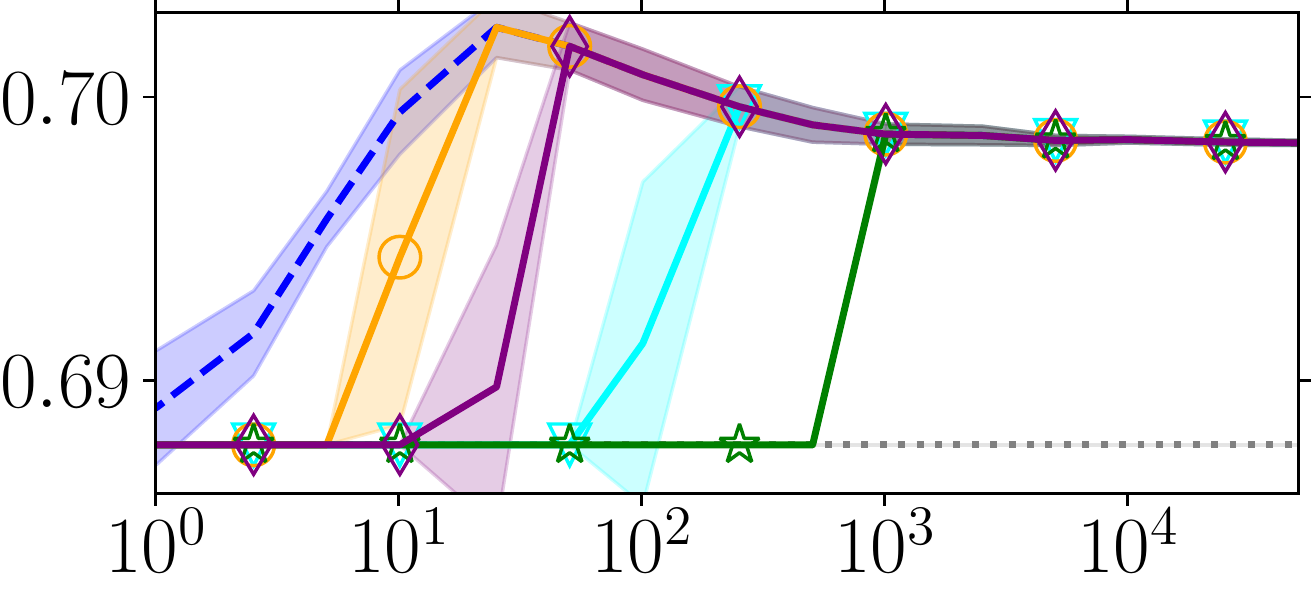}
\\
\cmidrule{2-4}
& \multicolumn{3}{c}{Clicks generated with $\alpha=0.025$.}
\\
\cmidrule{2-4}
\rotatebox[origin=lt]{90}{\hspace{0.5em} \small Train-NDCG} &
\includegraphics[scale=0.38]{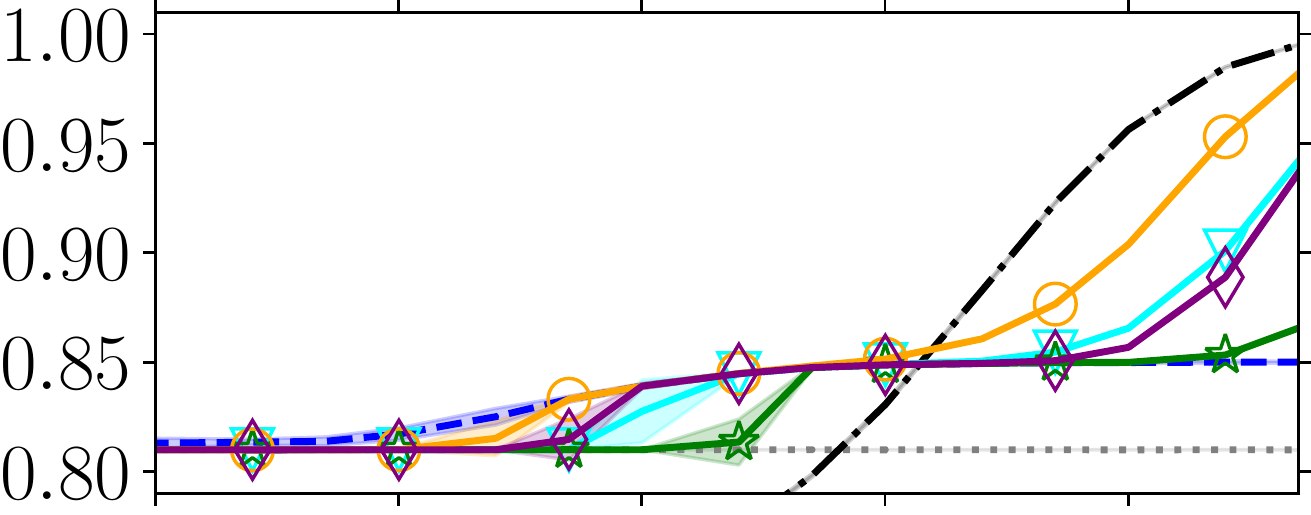} &
\includegraphics[scale=0.38]{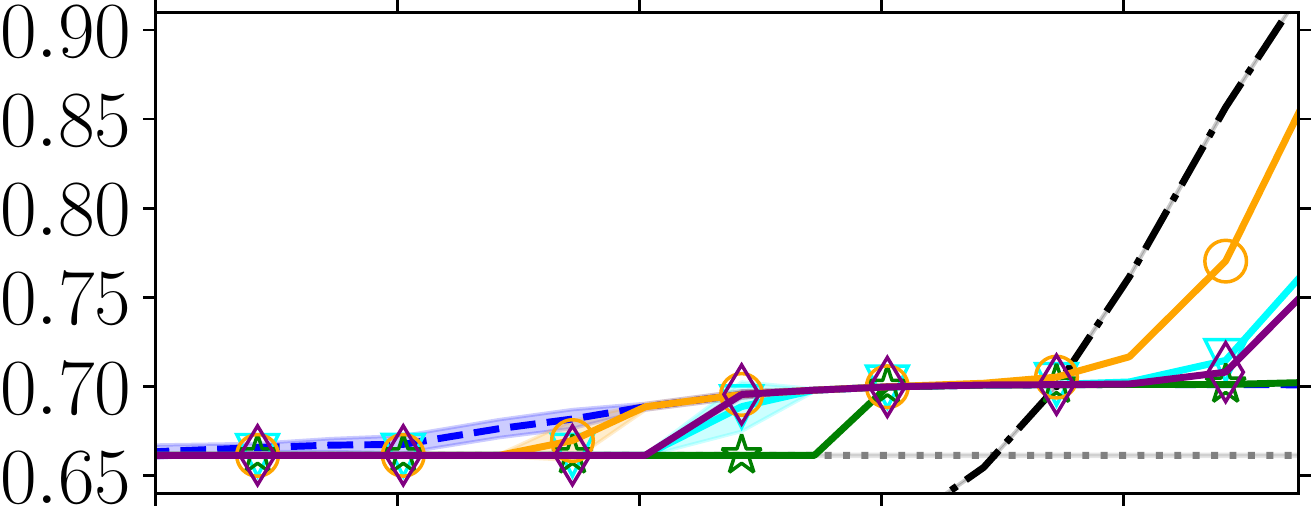} &
\includegraphics[scale=0.38]{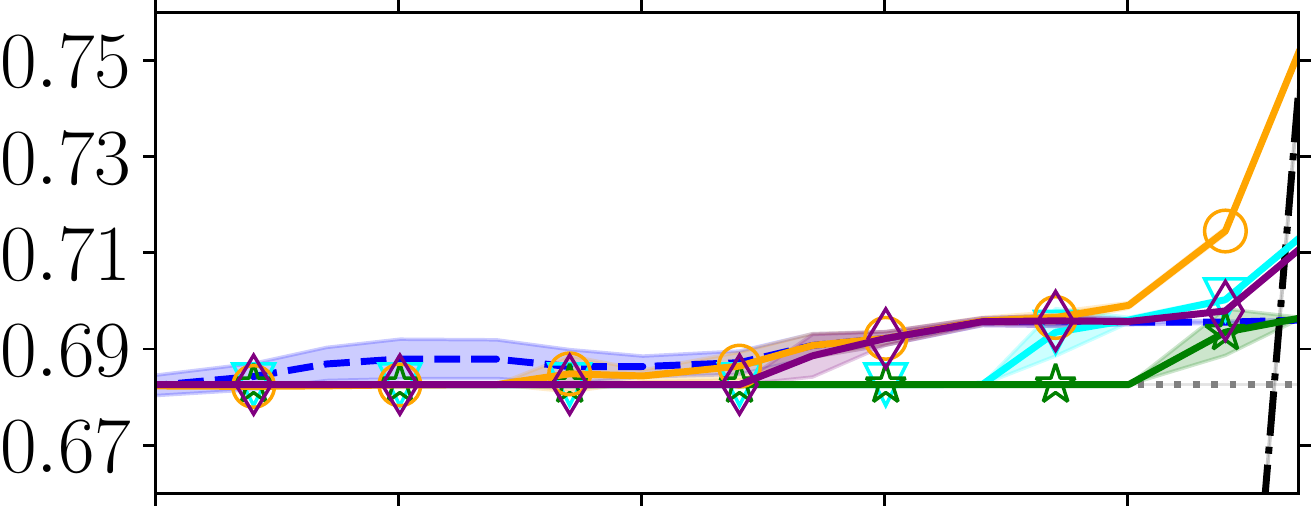} 
\\
\rotatebox[origin=lt]{90}{\hspace{1.5em} \small  Test-NDCG} &
\includegraphics[scale=0.38]{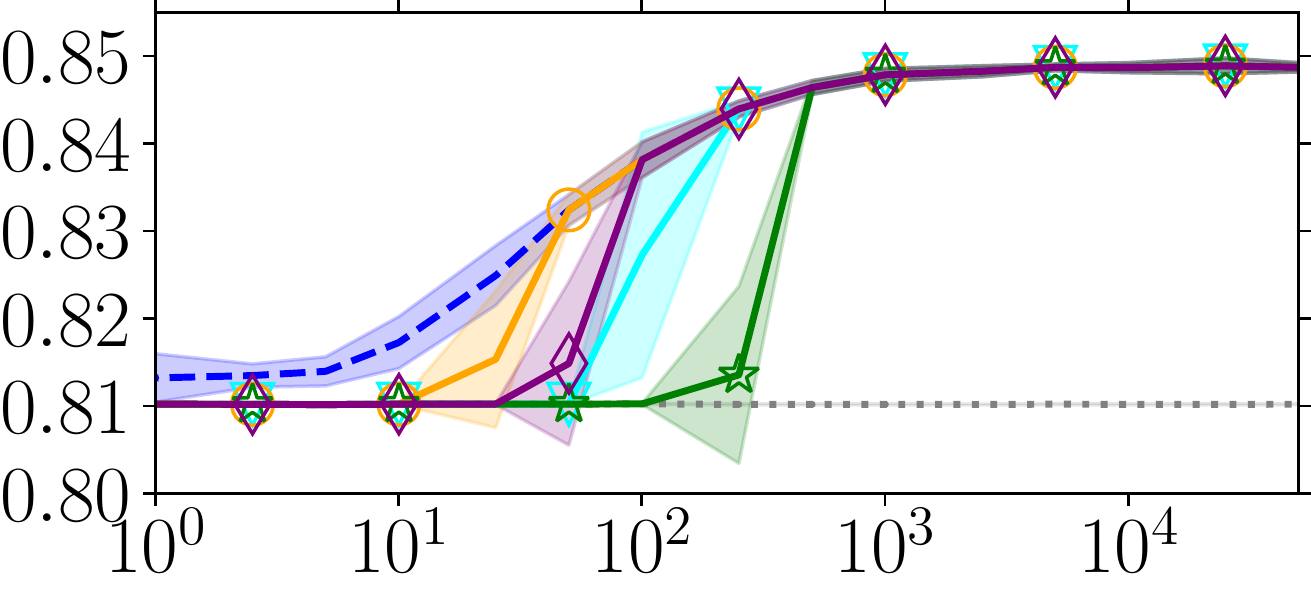} &
\includegraphics[scale=0.38]{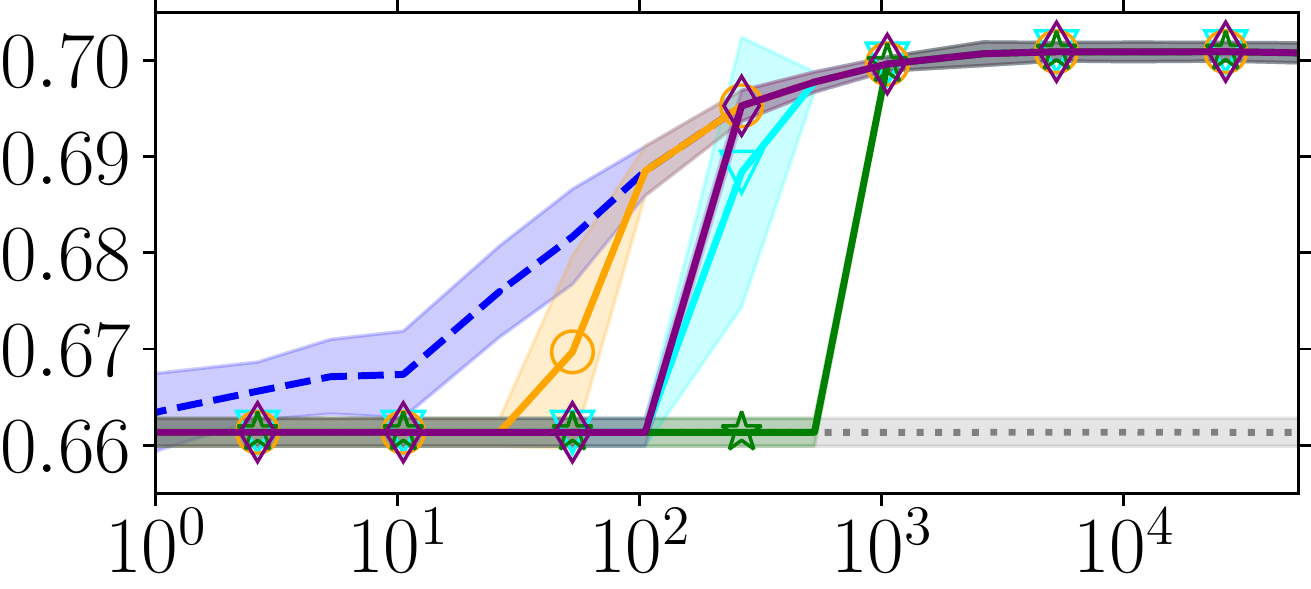} &
\includegraphics[scale=0.38]{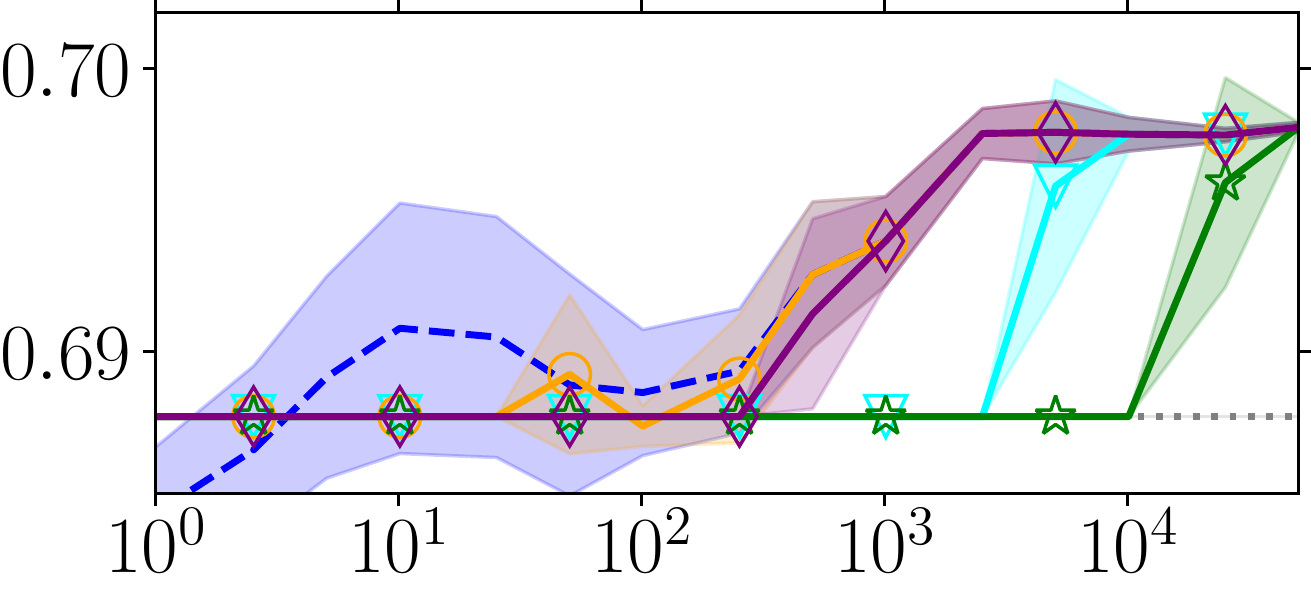}
\\
& \multicolumn{1}{c}{\small \hspace{0.5em} Mean Number of Clicks per Query}
& \multicolumn{1}{c}{\small \hspace{0.5em} Mean Number of Clicks per Query}
& \multicolumn{1}{c}{\small \hspace{0.5em} Mean Number of Clicks per Query}
\end{tabular}
\vspace{0.3\baselineskip}
\caption{
\ac{GENSPEC} compared to a meta-policy using the \acs{SEA} bounds~(see Section~\ref{sec:SEAresults}).
Notation is the same as in Figure~\ref{fig:mainresults}.
}
\label{fig:sea}
\end{figure*}
}

\section{Experimental Setup}
\label{sec:experimentalsetup}

In our experiments we compare \ac{GENSPEC} to:
\begin{enumerate*}[label=(\roman*)]
\item a single feature- based model and a single tabular model to evaluate if \ac{GENSPEC} truly safely combines the advantages of generalization and specialization;
\item the \ac{GENSPEC} strategy using the \ac{SEA} bounds, in order to test whether our novel bound on relative performance is truly more efficient; and lastly,
\item online bandit-style \ac{LTR} algorithms, to evaluate whether \ac{GENSPEC} provides the same high-performance convergence while avoiding initial periods of poor performance.
\end{enumerate*}
In order to perform reproducible experiments under varying circumstances, we make use of a semi-synthetic setup and evaluate both on previously seen and unseen queries.  

\subsection{The Semi-Synthetic Setup}

Our experimental setup is semi-synthetic: queries, relevance judgements, and documents come from industry datasets, while biased and noisy user interactions are simulated using probabilistic user models.
This setup is very common in the counterfactual and online \ac{LTR} literature~\citep{joachims2017unbiased, agarwal2019counterfactual, oosterhuis2019optimizing}.
We make use of the three largest \ac{LTR} industry datasets: \emph{Yahoo! Webscope}~\citep{Chapelle2011}, \emph{MSLR-WEB30k}~\citep{qin2013introducing}, and \emph{Istella}~\citep{dato2016fast}.
Each consists of a set of queries, with for each query a preselected set of documents; document-query combinations are only represented by feature vectors and a label indicating relevance according to expert annotators.
Labels range from $0$ (not relevant) to $4$ (perfectly relevant).
User issued queries are simulated by uniformly sampling from the training and validation partitions of the datasets.
Displayed rankings are generated by a logging ranker using a linear model optimized on $1\%$ of the training partition using supervised \ac{LTR}~\citep{joachims2017unbiased}.
Then, user examination is simulated with probabilities inverse to the displayed rank of a document:
$
P\big(O = 1 \mid d, y\big) = \frac{1}{\textit{rank}(d\mid y)}.
$
Finally, user clicks are generated according to the following formula using a single parameter $\alpha \in \mathbb{R}$:
\begin{equation}
P\big(C = 1 \mid o_i(d) = 1, r(q, d)\big) = 0.2 + \alpha \cdot \text{relevance\_label}(q, d).
\end{equation}
In our experiments, we use $\alpha = 0.2$ and $\alpha = 0.025$; the former represents an easier setting where relevance has a great effect on the click probability; the latter represents a more noisy and harder setting where relevance has a far smaller influence.
Clicks are only generated on the training and validation partitions; $50\%$ of the training clicks are separated for model selection~($\bounddata$); our feature-based models are linear models; hyperparameter optimization is done using counterfactual evaluation with clicks on the validation partition~\citep{joachims2017unbiased}.

Some of our baselines are online bandit algorithms; for these baselines no clicks are separated for $\bounddata$, and the algorithms are run online: clicks are not gathered using the logging policy but by applying the algorithms in an online interactive manner.

\subsection{Evaluation}

The evaluation metric we use is normalized \ac{DCG}~(Eq.~\ref{eq:dcg})~\citep{jarvelin2002cumulated} using the ground-truth labels from the datasets.
To evaluate the high-performance at convergence of tabular models, we do not apply a rank-cutoff when computing the metric; thus, an NDCG of $1.0$ indicates that \emph{all} documents are ranked perfectly.
Furthermore, we wish to evaluate the performance of \ac{GENSPEC} on queries with different frequencies.
Unfortunately, the datasets do not indicate how frequent each query is.
As a solution, we vary the number of total clicks from $100$ up to $10^9$ in total, uniformly spread over all queries.
We separately calculate performance on the test set (Test-NDCG) and the training set (Train-NDCG).
Because no clicks are ever generated on the test set, Test-NDCG shows the performance on previously unseen queries.
Metrics are always computed over all queries in the partition, thus every reported value of Train-NDCG is based on the entire training set, the same goes for Test-NDCG and the test set.
All reported results are averages over $10$ runs.

\section{Results and Discussion}

\subsection{Behavior of \ac{GENSPEC}}

First, we will contrast \ac{GENSPEC} with a single feature-based model and a single tabular model.
Figure~\ref{fig:mainresults} shows the performance of 
\begin{enumerate*}[label=(\roman*)]
\item \ac{GENSPEC} with different levels of confidence for its bounds ($\epsilon$),
along with that of 
\item the logging policy, 
\item the feature-based model, and 
\item the tabular model between which the \ac{GENSPEC} chooses.
\end{enumerate*}
We see that the feature-based model requires few clicks to improve over the logging policy but is not able to reach optimal levels of performance.
The performance of the tabular model, on the other hand, is initially far below the logging policy.
However, after enough clicks have been gathered, performance increases until the optimal ranking is found; when click noise is limited ($\alpha = 0.2$) it reaches perfect performance on all three datasets (Train-NDCG).
On the unseen queries where there are no clicks (Test-NDCG), the tabular model is unable to learn anything and provides random performance (not displayed in Figure~\ref{fig:mainresults}).
The initial period of poor performance can be very detrimental to queries that do not receive a large number of clicks.
Prior work has found that web-search queries follow a long-tail distribution~\citep{silverstein1999analysis, spink2002us}; \citet{white2007studying} found that 97\% of queries received fewer than $10$ clicks over six months.
For such queries, users may only experience the initial poor performance of the tabular model, and never see the improvements it brings at convergence.
This possibility can be a large deterrent from applying tabular models in practice~\citep{wu2016conservative}.
Furthermore, our results indicate there is no simple way to determine when the tabular model is the best choice, i.e., depending on the dataset and the level of noise, this could be after $10^2$, $10^3$ or more than $10^4$ clicks.
Therefore, it is unlikely that a simple heuristic can accurately detect these moments in practice.

Finally, we see that by choosing between the three models GEN\-SPEC combines properties of all: after a few clicks it deploys the feature-based model and thus outperforms the logging policy; as more clicks are gathered, the tabular model is activated on queries, further improving performance.
With $\alpha = 0.2$ \ac{GENSPEC} with $\epsilon \leq 0.75$ reaches perfect Train-NDCG performance on all three datasets by widely deploying the tabular model.
However, unlike the tabular model, the performance of \ac{GENSPEC} (with $\epsilon > 0$) never drops below the logging policy.
Moreover, we never observe a situation where an increase in the number of clicks results in a decrease in the mean performance of \ac{GENSPEC}.
As expected, there is a delay between when the tabular model is the optimal choice and when \ac{GENSPEC} deploys the tabular model on queries.
Thus, while the usage of confidence bounds prevents the performance from dropping below the level of the logging policy, it does so at the cost of this delay.
When \ac{GENSPEC} does not use any bounds, it deploys the tabular model earlier; in some cases these deployments result in worse performance than the logging policy, albeit less than the tabular model on its own.
In all our observed results, a confidence of $\epsilon = 0.01$ was enough to prevent any decreases in performance.

To conclude, our experimental results show that \ac{GENSPEC} combines the high-performance at convergence of specialization and the safe robustness of generalization.
In contrast to tabular models, which results in very poor performance when not enough clicks have been gathered, \ac{GENSPEC} effectively avoids incorrect deployment and under our tested conditions it never performs worse than the logging policy.
Meanwhile, \ac{GENSPEC} achieves considerable gains in performance at convergence, in contrast with feature-based models.
We only observe a very small delay between when the feature-based model is the optimal choice and when \ac{GENSPEC} deploys it.
We conclude that \ac{GENSPEC} is generally preferable to pure feature-based counterfactual \ac{LTR}.
Compared to pure tabular counterfactual \ac{LTR}, \ac{GENSPEC} is the best choice in situations where periods of poor performance should be avoided~\citep{wu2016conservative} or when not all queries receive large numbers of clicks~\citep{white2007studying}.

\vspace*{-0.75\baselineskip}
\subsection{Effectiveness of Relative Bounding}
\label{sec:SEAresults}

To evaluate the efficiency of our relative performance bounds, we apply the \ac{SEA} bounds due to~\citet{jagerman2020safety} to the \ac{GENSPEC} strategy.
This means that two bounds are used for every choice between two models, each bounding the performance of an individual model.
For a fair comparison, we adapt \ac{SEA} to choose between the same models as \ac{GENSPEC} and provide it with the same click data.

Figure~\ref{fig:sea} displays the results of this comparison.
Across all settings, \ac{GENSPEC} deploys models much earlier than \ac{SEA} with the same level of confidence.
While they converge at the same levels of performance, \ac{GENSPEC} requires considerably less data, e.g., on the Istella dataset with $\alpha=0.025$, \ac{GENSPEC} deploys models with $10$ times less data.
Thus, we conclude that the relative bounds of \ac{GENSPEC} are much more efficient than the existing bounding approach of \ac{SEA} (confirming the theory in Appendix~\ref{sec:theoryrelativebounds}).

\vspace*{-0.5\baselineskip}
\subsection{Comparison to Online \ac{LTR} Bandits}
\label{sec:onlineltr}

Last, we compared \ac{GENSPEC} to online bandit \ac{LTR} algorithms~\citep{kveton2015cascading, katariya2016dcm}.
Unlike counterfactual \ac{LTR}, these bandit methods learn using online interventions: at each timestep they choose which ranking to display to users.
As baselines we use the Hotfix algorithm~\citep{zoghi2016click} and the \ac{PBM}~\citep{lagree2016multiple}.
The Hotfix algorithm is a very general approach, it completely randomly shuffles the top-$n$ items and ranks them based on pairwise preferences inferred from clicks.
The main downside of the Hotfix approach is that its randomization is very detrimental to the user experience.
In our results, we only report the performance of the ranking produced by the Hotfix baseline, not of the randomized rankings used to gather clicks.
We apply two versions of the Hotfix algorithm, one for top-10 reranking and another for the complete ranking.
\ac{PBM} is perfectly suited for our task as it makes the same assumptions about user behavior as our experimental setting.
We apply PMB-PIE~\citep{lagree2016multiple}, which results in PBM always displaying the ranking it expects to perform best, thus attempting to maximize the user experience during learning.
These methods all optimize a tabular ranking model: the bandit baselines memorize the best rankings and do not depend on features at all.
Consequently, their learned policies cannot be applied to previously unseen queries, hence, we do not report their Test-NDCG.
 
Figure~\ref{fig:bandits} displays the results for this comparison.
We see that when $\alpha = 0.2$ Hotfix-Complete, \ac{PBM} and \ac{GENSPEC} all reach perfect Train-NCDG; however, Hotfix-Complete and \ac{PBM} reach convergence much earlier than \ac{GENSPEC}.
We attribute this difference to two reasons:
\begin{enumerate*}[label=(\roman*)]
\item the online interventions of the bandit baselines, and
\item the delay in deployment added by \ac{GENSPEC}'s usage of confidence bounds.
\end{enumerate*}
Similar to the tabular model, the earlier moment of convergence of the bandit baselines comes at the cost of an initial period of very poor performance.
We conclude that if only the moment of reaching optimal performance matters, \ac{PBM} is the best choice of method.
However, if periods of poor performance should be avoided~\citep{wu2016conservative}, or if some queries may not receive large numbers of clicks~\citep{white2007studying}, \ac{GENSPEC} is the better choice.
An additional advantage is that \ac{GENSPEC} is a counterfactual method and does not have to be applied online like the bandit baselines.
Overall \ac{GENSPEC} is thus the safest choice w.r.t. the user experience since it avoids both online interventions and initial periods of poor performance.

{\renewcommand{\arraystretch}{0.63}
\begin{figure*}[t]
\centering
\begin{tabular}{c r r r r}
&
 \multicolumn{1}{c}{  Yahoo! Webscope}
&
 \multicolumn{1}{c}{  MSLR-WEB30k}
&
 \multicolumn{1}{c}{  Istella}
 &
\multirow{6}{*}[-6.5em]{
\includegraphics[scale=.33]{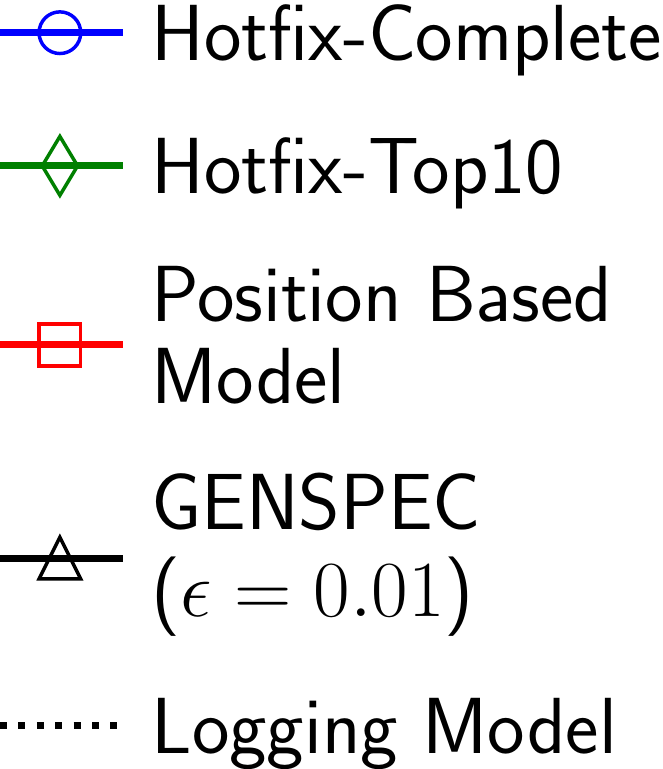}
} 
\\
\cmidrule{2-4}
& \multicolumn{3}{c}{Clicks generated with $\alpha=0.2$.}
\\
\cmidrule{2-4}
\rotatebox[origin=lt]{90}{\hspace{0.1em} \small Train-NDCG} &
\includegraphics[scale=0.38]{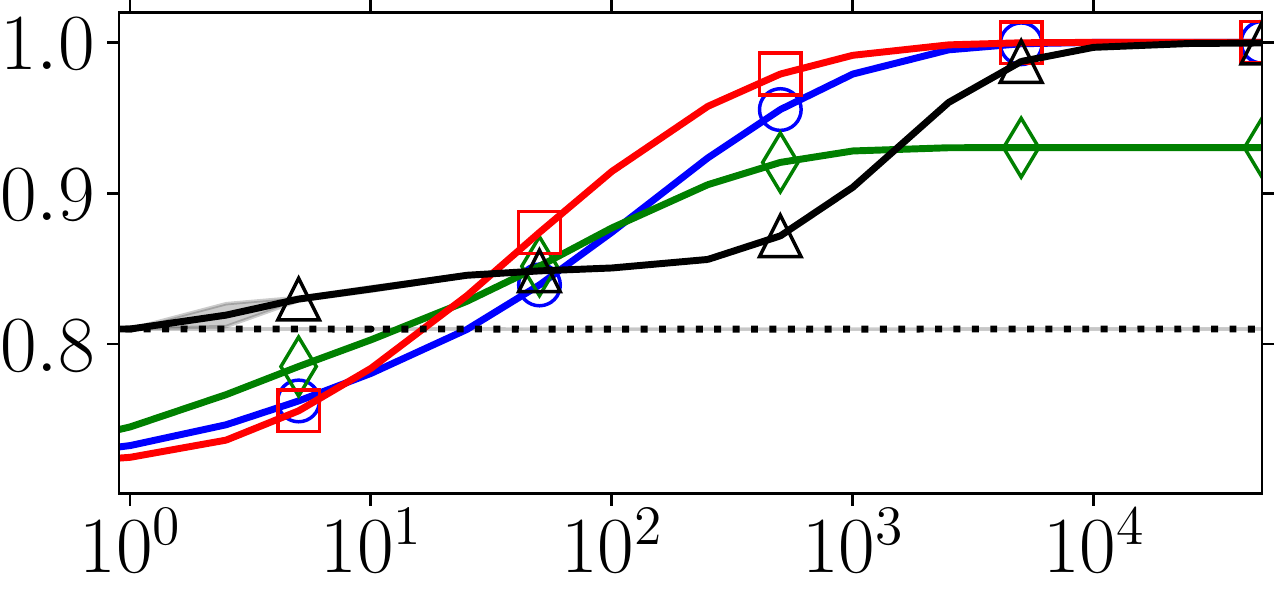} &
\includegraphics[scale=0.38]{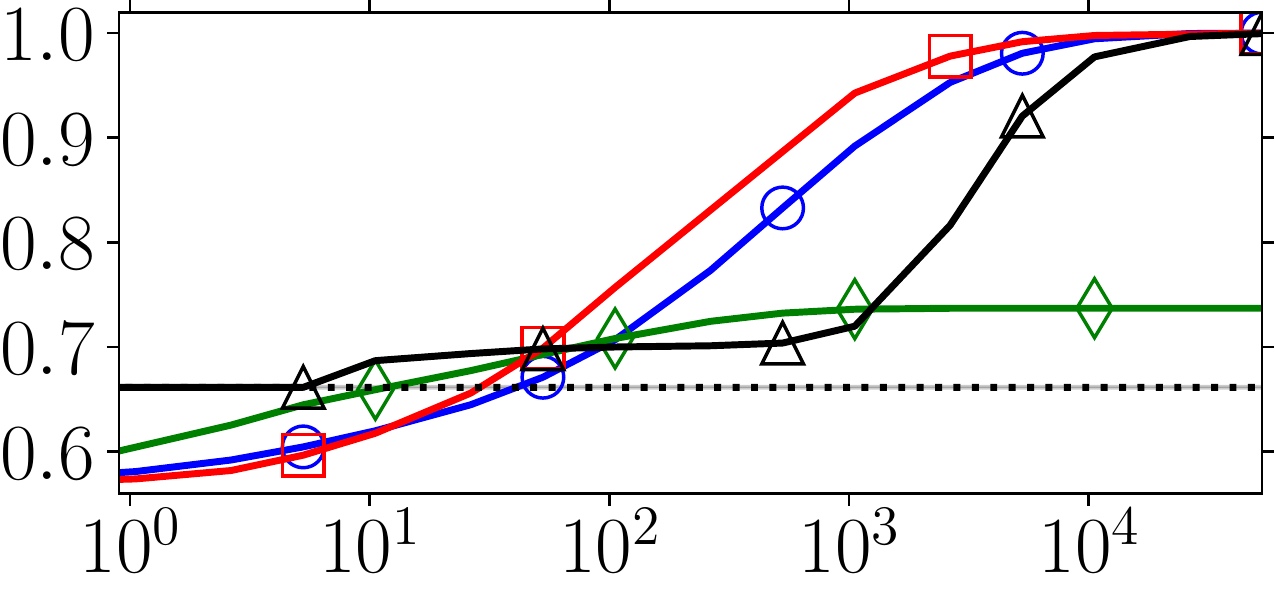} &
\includegraphics[scale=0.38]{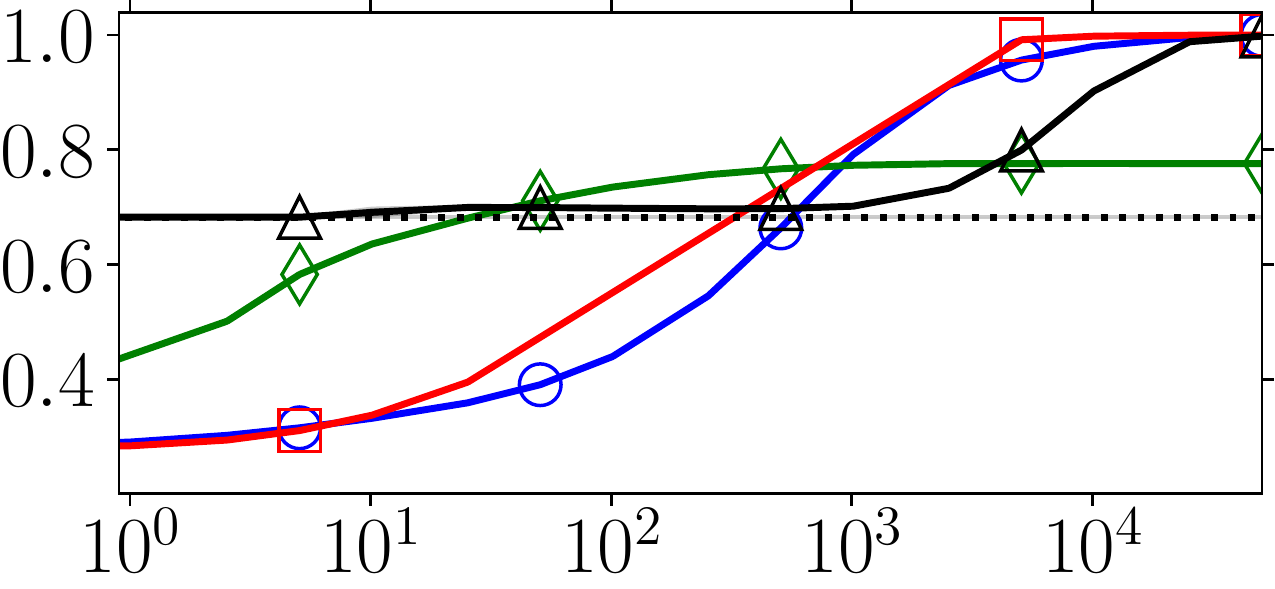} 
\\
\cmidrule{2-4}
& \multicolumn{3}{c}{Clicks generated with $\alpha=0.025$.}
\\
\cmidrule{2-4}
\rotatebox[origin=lt]{90}{\hspace{0.7em} \small Train-NDCG} &
\includegraphics[scale=0.38]{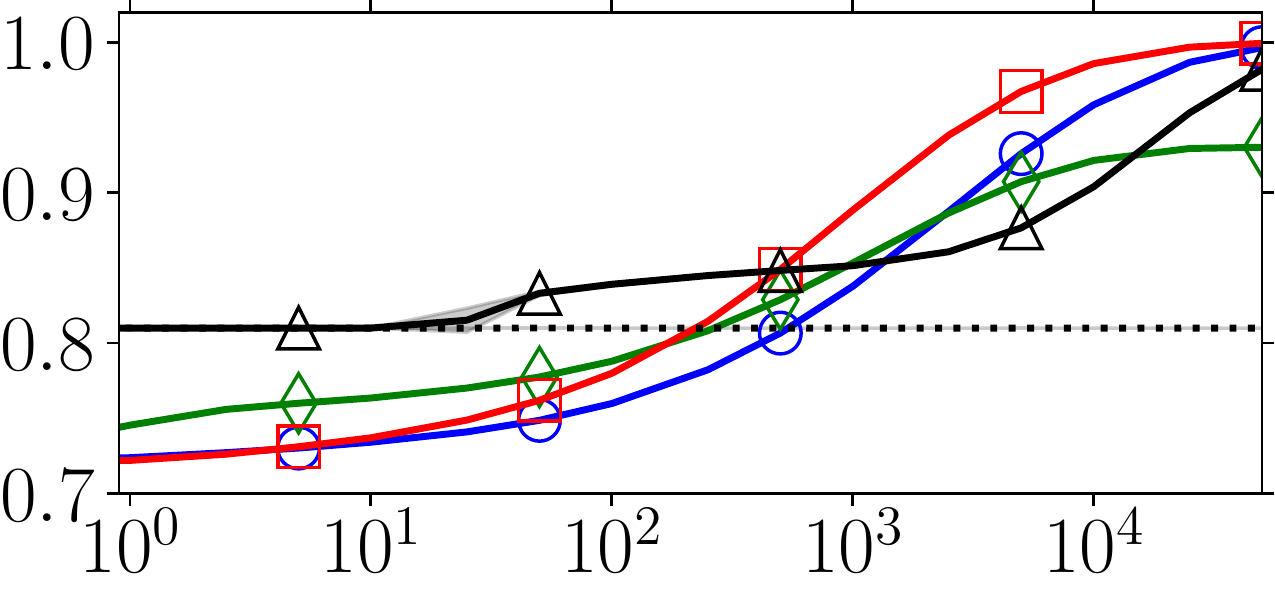} &
\includegraphics[scale=0.38]{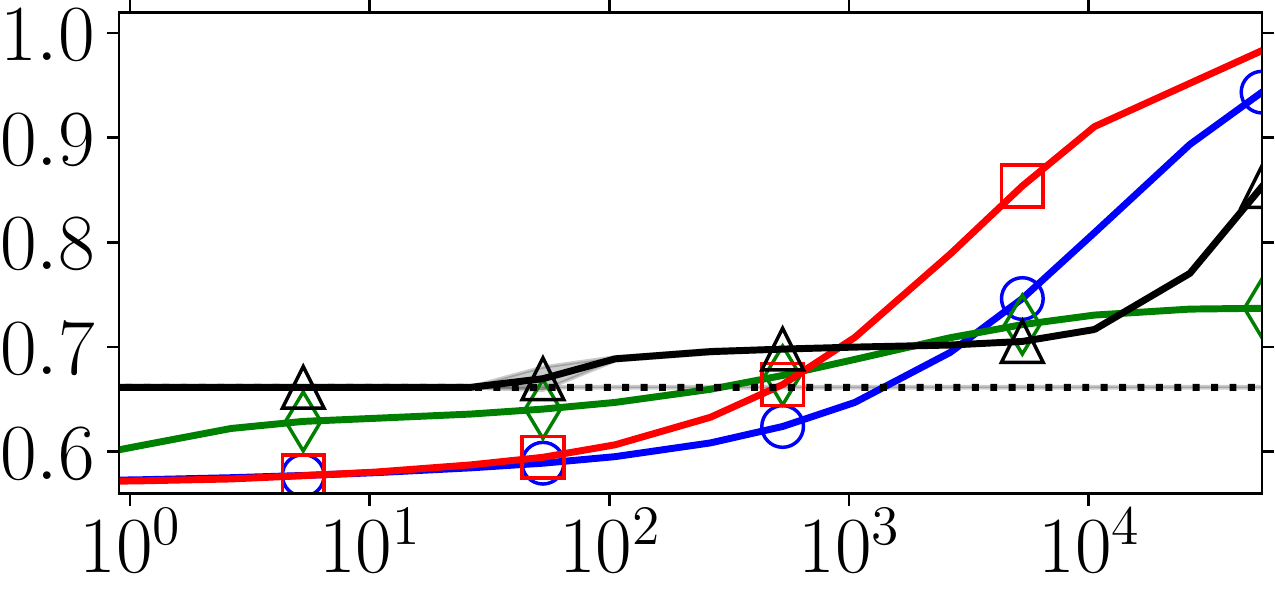} &
\includegraphics[scale=0.38]{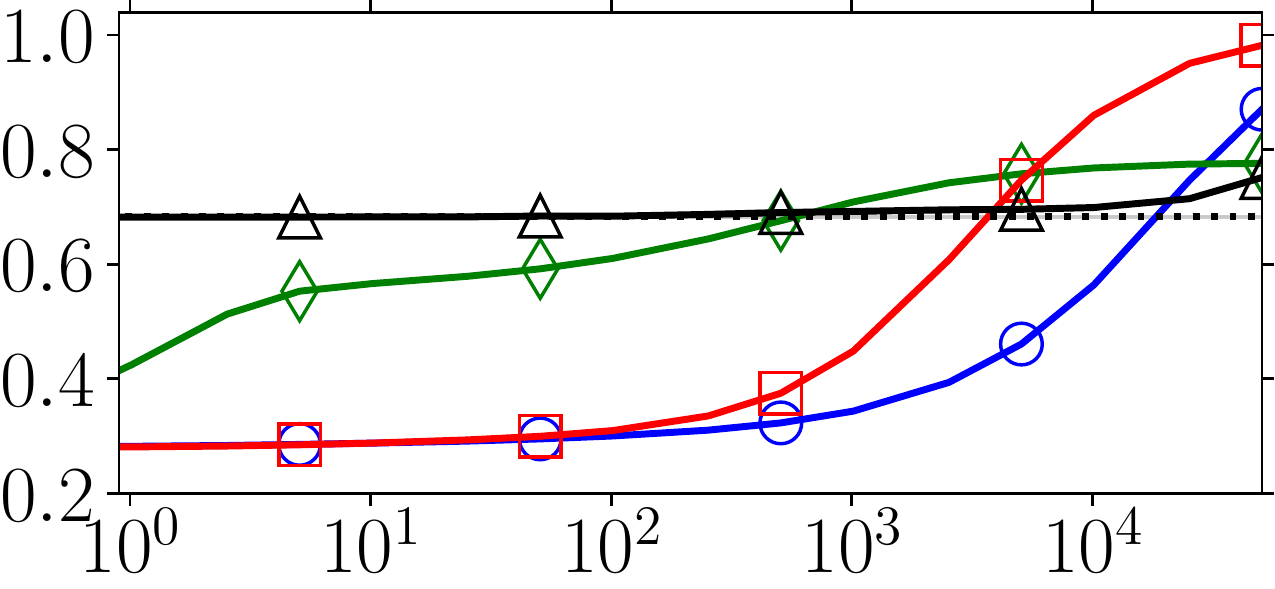} 
\\
& \multicolumn{1}{c}{\small \hspace{0.5em} Mean Number of Clicks per Query}
& \multicolumn{1}{c}{\small \hspace{0.5em} Mean Number of Clicks per Query}
& \multicolumn{1}{c}{\small \hspace{0.5em} Mean Number of Clicks per Query}
\end{tabular}
\vspace{0.3\baselineskip}
\caption{
\ac{GENSPEC} compared to various online \ac{LTR} bandits (see Section~\ref{sec:onlineltr}).
Notation is the same as for Figure~\ref{fig:mainresults}.
}
\label{fig:bandits}
\vspace{-1.1\baselineskip}
\end{figure*}
}

\section{Conclusion}
\label{sec:Conclusion}

In this paper we have introduced the \acf{GENSPEC} framework for safe query specialization in counterfactual \ac{LTR}.
It simultaneously learns a feature-based model to perform well across all queries, and a tabular model consisting of many memorized rankings optimized for individual queries.
Per query, \ac{GENSPEC} uses high-confidence bounds to choose between deploying the logging policy, the feature-based model, or the tabular model.
Our results show that \ac{GENSPEC} combines the high performance of a successfully specialized tabular model on queries with sufficiently many interactions, with the safe robust performance of a feature-based model on queries that were previously unseen or where little data is available.
As a result, it avoids the low performance at convergence of feature-based models, and the initial poor performance of the tabular models.

We expect \ac{GENSPEC} to be a framework that will be used for other types of specialization in the future.
For instance, we think personalization for \ac{LTR} is a promising application.
Moreover, in Appendix~\ref{sec:genspeccontextualbandit} we describe how \ac{GENSPEC} could be applied to contextual bandit problems outside of \ac{LTR}.
There are many fruitful directions future work could explore with the \ac{GENSPEC} framework.

\vspace{-0.25\baselineskip}
\section*{Reproducibility}
For full reproducibility this work only made use of publicly available data and our complete experimental implementation is publicly available at \url{https://github.com/HarrieO/2021WWW-GENSPEC}.

\vspace{-0.25\baselineskip}
\begin{acks}
This research was supported by the Netherlands Organisation for Scientific Research (NWO)
under pro\-ject nr 612.\-001.\-551.
All content represents the opinion of the authors, which is not necessarily shared or endorsed by their respective employers and/or sponsors.
\end{acks}

\vspace{-0.25\baselineskip}
\bibliographystyle{ACM-Reference-Format}
\bibliography{references}

\appendix

\section{Proof of Unbiasedness for Counterfactual Learning to Rank}
\label{sec:counterfactualproof}

In this section we will prove that the \ac{IPS} estimate $\hat{\mathcal{R}}$~(Eq.~\ref{eq:rewardestimate}) can be used to unbiasedly optimize the true reward $\mathcal{R}$~(Eq.~\ref{eq:truereward}), as claimed in Section~\ref{sec:counterfactualltr}.
First, we consider the expected value for an observed click $c_i(d)$ using Eq.~\ref{eq:observed}:
\begin{align}
&\mathds{E}_{y_i,o_i}\big[c_i(d)\big] 
= \mathds{E}_{y_i}\Big[P\big(C = 1 | o_i(d) = 1, r(q_i, d)\big) \cdot P\big(O=1 | d, y_i\big)\Big]  \nonumber\\
&= P\big(C = 1 \,|\, o_i(d) = 1, r(q_i,d)\big) \cdot \Bigg(\sum_{y \in \pi_0}  P\big(O=1 \,|\ d, y\big) \cdot \pi_0(y \,|\ q_i) \Bigg)  \nonumber \\
&= \rho_i(d) \cdot P\big(C = 1 \mid o_i(d) = 1, r(q_i,d)\big). 
\end{align}
Note that $y_i$ is the ranking displayed at interaction $i$ while $y$ is the ranking being evaluated,
the expected value for the \ac{IPS} estimator:
\begin{equation}
\begin{split}
\mathds{E}_{o_i,y_i}&\big[\hat{\Delta}(y \mid c_i, \rho_i)\big]
= \mathds{E}_{o_i,y_i}\Bigg[\sum_{d \in y }\lambda\big(\textit{rank}(d \mid y)\big) \cdot \frac{c_i(d)}{\rho_i(d)} \Bigg] \\ 
& = \sum_{d \in y} \lambda\big(\textit{rank}(d \,|\, y)\big) \cdot P\big(C = 1 \,|\, o_i(d) = 1, r(q_i, d)\big). 
\end{split}
\end{equation}
This step assumes that $\rho_i(d) > 0$, i.e., that every item has a non-zero probability of being examined~\citep{joachims2017unbiased,oosterhuis2020topkrankings}.
While $\mathds{E}_{o_i,y_i}[\hat{\Delta}(y \mid c_i, \rho_i)]$ and $\Delta(y \mid q_i, r)$ are not necessarily equal, using Eq.~\ref{eq:observed} we see that they are proportional with some offset $C$:
\begin{align}
\mathds{E}_{o_i,y_i}\big[\hat{\Delta}(y \mid c_i, \rho_i)\big] &\propto
\Big(\sum_{d \in y}\lambda\big(\textit{rank}(d \mid y)\big) \cdot r(q_i, d) \Big) + C \nonumber \\
&= \Delta(y \mid q_i, r) + C,
\end{align}
where $C$ is a constant: $C = \big(\sum^K_{i=1}\lambda(i)\big) \cdot \mu$.
Therefore, in expectation $\hat{\mathcal{R}}$ and $\mathcal{R}$ are also proportional with the same constant offset:
\begin{equation}
\mathds{E}_{o_i,y_i}\big[\hat{\mathcal{R}}(\pi \mid \mathcal{D})\big] \propto \mathcal{R}(\pi) + C.
\end{equation}
Consequently, the estimator can be used to unbiasedly estimate the preference between two models:
\begin{equation}
\mbox{}\hspace*{-2mm}
\mathds{E}_{o_i,y_i}\big[\hat{\mathcal{R}}(\pi_1 \mid \mathcal{D})\big] < \mathds{E}_{o_i,y_i}\big[\hat{\mathcal{R}}(\pi_2 \mid \mathcal{D})\big]
\Leftrightarrow \mathcal{R}(\pi_1) < \mathcal{R}(\pi_2).
\hspace*{-2mm}\mbox{}
 \label{eq:relativeproof}
\end{equation}
Moreover, this implies that maximizing the estimated performance unbiasedly optimizes the actual reward:
\begin{equation}
 \argmax_{\pi} \mathds{E}_{o_i,y_i}\big[\hat{\mathcal{R}}(\pi \mid \mathcal{D})\big] = \argmax_{\pi} \mathcal{R}(\pi).
 \label{eq:maxproof}
\end{equation}
This concludes the proof; we have shown that $\hat{\mathcal{R}}$ is suitable for unbiased \ac{LTR}, since  it can be used to find the optimal model.

\section{Efficiency of Relative Bounding}
\label{sec:theoryrelativebounds}

In this section, we prove that the relative bounds of \ac{GENSPEC} are more efficient than \ac{SEA} bounds~\citep{jagerman2020safety}, when the covariance between the reward estimates of two models is positive:
\begin{equation}
\text{cov}\big(\hat{\mathcal{R}}(\pi_1 \mid \mathcal{D}), \hat{\mathcal{R}}(\pi_2 \mid \mathcal{D})\big) > 0.
\end{equation}
This means that \ac{GENSPEC} will deploy a model earlier than \ac{SEA} if there is positive covariance; since both estimates are based on the same interaction data $\mathcal{D}$, a high covariance is very likely.

Let us first consider when \ac{GENSPEC} deploys a model.
Deployment by \ac{GENSPEC} depends on whether a relative confidence bound is greater than the estimated difference in performance~(cf.\ Eq.~\ref{eq:genspecchoice1} and ~\ref{eq:genspecchoice2}).
For two models $\pi_1$ and $\pi_2$ deployment happens when:
\begin{equation}
\hat{\mathcal{R}}(\pi_1 \mid \mathcal{D}) - \hat{\mathcal{R}}(\pi_2 \mid \mathcal{D}) - \textit{CB}(\pi_1, \pi_2 \mid \mathcal{D})  > 0.
\end{equation}
Thus the bound has to be smaller than the estimated performance difference:
\begin{equation}
\textit{CB}(\pi_1, \pi_2 \mid \mathcal{D}) <
\hat{\mathcal{R}}(\pi_1 \mid \mathcal{D}) - \hat{\mathcal{R}}(\pi_2 \mid \mathcal{D}).
\label{eq:relativeboundreq}
\end{equation}
In contrast, \ac{SEA} does not use a single bound, but two bounds on the individual performances of the models.
For clarity, we describe the \ac{SEA} bound in our notation.
First, we have $R^{\pi_j}_{i,d}$, the observed reward for a $d$ at interaction $i$ for model $\pi_j$:
\begin{equation}
R^{\pi_j}_{i,d} =  \frac{c_i(d)}{\rho_i(d)} \sum_{y \in \pi_j} \pi_j(y \mid q_i)
 \cdot  \lambda\big(\textit{rank}(d \mid y)\big).
\end{equation}
Then we have a $\nu^{\pi_j}$ for each model:
\begin{align}
\nu^{\pi_j} =  \frac{ 2  |\mathcal{D}|  K  \ln\big(\frac{2}{1-\epsilon}\big)}{|\mathcal{D}| K-1}  \sum_{i \in \mathcal{D}} \sum_{d \in y_i} \big(K \cdot R^{\pi_j}_{i,d} -  \hat{\mathcal{R}}(\pi_j \mid \mathcal{D})  \big)^2,
\nonumber
\end{align}
which we use in the confidence bound for a single model $\pi_j$:
\begin{align}
\textit{CB}
(\pi_j \mid \mathcal{D})
= \frac{7 K b\ln\big(\frac{2}{1-\epsilon}\big)}{3(|\mathcal{D}|  K-1)} + \frac{1}{|\mathcal{D}|  K}  
 \cdot \sqrt{\nu^{\pi_j}}.
\end{align}
We note that the $b$ parameter has the same value for both the relative and single confidence bounds.
\ac{SEA} chooses between model by comparing their upper and lower confidence bounds:
\begin{equation}
\hat{\mathcal{R}}(\pi_1 \mid \mathcal{D}) - \textit{CB}(\pi_1 \mid \mathcal{D}) > \hat{\mathcal{R}}(\pi_2 \mid \mathcal{D}) + \textit{CB}(\pi_2 \mid \mathcal{D}).
\end{equation}
In this case, the summation of the bounds has to be smaller than the estimated performance difference:
\begin{equation}
\textit{CB}(\pi_1 \mid \mathcal{D})  + \textit{CB}(\pi_2 \mid \mathcal{D})  < \hat{\mathcal{R}}(\pi_1 \mid \mathcal{D}) - \hat{\mathcal{R}}(\pi_2 \mid \mathcal{D}).
\label{eq:singleboundreq}
\end{equation}
We can now formally describe under which condition \ac{GENSPEC} is more efficient than \ac{SEA}:
by combining Eq.~\ref{eq:relativeboundreq} and Eq.~\ref{eq:singleboundreq}, we see that relative bounding is more efficient when:
\begin{equation}
\textit{CB}(\pi_1, \pi_2 \mid \mathcal{D}) <
\textit{CB}(\pi_1 \mid \mathcal{D})  + \textit{CB}(\pi_2 \mid \mathcal{D}).
\end{equation}
We notice that $\mathcal{D}$, $K$, $b$ and $\epsilon$ have the same value for both confidence bounds, thus we only require:
\begin{align}
\sqrt{\nu} < \sqrt{\nu^{\pi_1}} + \sqrt{\nu^{\pi_2}}.
\end{align}
If we assume that $\mathcal{D}$ is sufficiently large, we see that $\sqrt{\nu}$ approximates the standard deviation scaled by some constant:
\begin{align}
\sqrt{\nu} \approx C \cdot \sqrt{\text{var}\big(\hat{\delta}(\pi_1, \pi_2 | \mathcal{D})\big)}, \text{with } C = \sqrt{\frac{ 2  |\mathcal{D}|^2  K^2  \ln\big(\frac{2}{1-\epsilon}\big)}{|\mathcal{D}| K-1}}.
\end{align}
Since the bounds prevent deployment until enough certainty has been gained, we think it is safe to assume that $\mathcal{D}$ is large enough for this approximation before any deployment takes place. 

To keep our notation concise, we use:
$\hat{\delta} = \hat{\delta}(\pi_1, \pi_2 \mid \mathcal{D})$,
$\hat{\mathcal{R}}_1 = \hat{\mathcal{R}}(\pi_1 \mid \mathcal{D})$,
and
$\hat{\mathcal{R}}_2 = \hat{\mathcal{R}}(\pi_2 \mid \mathcal{D})$.
Using the same approximations for $\sqrt{\nu^{\pi_1}}$ and $\sqrt{\nu^{\pi_2}}$ we get:
\begin{align}
\sqrt{\text{var}(\hat{\delta})} < \sqrt{\text{var}(\hat{\mathcal{R}}_1)} + \sqrt{\text{var}(\hat{\mathcal{R}}_2)}.
\end{align}
By making use of the Cauchy-Schwarz inequality, we can derive the following lower bound:
\begin{align}
\sqrt{\text{var}(\hat{\mathcal{R}}_1) + \text{var}(\hat{\mathcal{R}}_2)}
\leq \sqrt{\text{var}(\hat{\mathcal{R}}_1)} + \sqrt{\text{var}(\hat{\mathcal{R}}_2)}.
\end{align}
Therefore, the relative bounding of \ac{GENSPEC} must be more efficient when the following is true:
\begin{align}
\text{var}(\hat{\delta})
< \text{var}(\hat{\mathcal{R}}_1) + \text{var}(\hat{\mathcal{R}}_2),
\end{align}
i.e., the variance of the relative estimator must be less than the sum of the variances of the estimators for the individual model.
Finally, by rewriting $\text{var}(\hat{\delta})$ to:
\begin{equation}
\mbox{}\hspace*{-2mm}
\text{var}(\hat{\delta})
=
\text{var}(\hat{\mathcal{R}}_1 - \hat{\mathcal{R}}_2)
= \text{var}(\hat{\mathcal{R}}_1) + \text{var}(\hat{\mathcal{R}}_2) - 2\text{cov}(\hat{\mathcal{R}}_1, \hat{\mathcal{R}}_2),
\hspace*{-2mm}\mbox{}
\end{equation}
we see that the relative bounds of \ac{GENSPEC} are more efficient than the multiple bounds of \ac{SEA}
if the covariance between $\hat{\mathcal{R}}_1$ and $\hat{\mathcal{R}}_2$ is positive:
\begin{equation}
\text{cov}(\hat{\mathcal{R}}_1, \hat{\mathcal{R}}_2) > 0.
\end{equation}
Remember that both estimates are based on the same interaction data: $\hat{\mathcal{R}}_1 = \hat{\mathcal{R}}(\pi_1 | \mathcal{D})$,
and
$\hat{\mathcal{R}}_2 = \hat{\mathcal{R}}(\pi_2 | \mathcal{D})$.
Therefore, they are based on the same clicks and propensities scores, thus it is extremely likely that the covariance between the estimates is positive.
Correspondingly, it is also extremely likely that the relative bounds of \ac{GENSPEC} are more efficient than the bounds used by \ac{SEA}.

\section{GENSPEC for Contextual Bandits}
\label{sec:genspeccontextualbandit}

So far we have discussed \ac{GENSPEC} for counterfactual \ac{LTR}. 
We will now show that it is also applicable to the broader contextual bandit problem.
Instead of a query $q$, we now keep track of an arbitrary context $z \in \{1,2,\ldots\}$ where
$
z_i \sim P(Z).
$
Data is gathered using the logging policy $\pi_0$:
$
a_i \sim \pi_0(a \mid z_i),
$
where $a$ indicates an action.
However, unlike the \ac{LTR} case, the rewards $r_i$ are observed directly:
$
r_i \sim P(r \mid a_i, z_i).
$
With the propensities
$
\rho_i = \pi_0(a_i \mid z_i)
$
the data is:
$
\mathcal{D} = \big \{(r_i, a_i, \rho_i, z_i)\big \}^N_{i=1};
$
for specialization the data is filtered per context $z$:
$
\mathcal{D}_z = \big \{ (r_i, a_i, \rho_i,  z_i) \in \mathcal{D}\mid z_i = z \big\}.
$
Again, data for training $\traindata$ and for model selection $\bounddata$ are separated.
The reward is estimated with an \ac{IPS} estimator:
\begin{equation}
\hat{\mathcal{R}}(\pi \mid \mathcal{D}) = \frac{1}{|\mathcal{D}|} \sum_{i \in \mathcal{D}} \frac{r_i}{\rho_i} \, \pi(a_i \mid z_i).
\end{equation}
Again, we have a policy trained for generalization $\pi_g$ and another for specialization $\pi_z$, and the copies trained on $\traindata$: $\pi_g'$ and $\pi_z$'.
The difference between two policies is estimated by:
$
\hat{\delta}(\pi_1, \pi_2 \mid \mathcal{D}) = \hat{\mathcal{R}}(\pi_1 \mid \mathcal{D}) - \hat{\mathcal{R}}(\pi_2 \mid \mathcal{D}).
$
We differ from the \ac{LTR} approach by estimating the bounds using:
\begin{equation}
R_i =  \frac{r_i}{\rho_i} \, \big(\pi_1(a_i \mid x_i, z_i) - \pi_2(a_i \mid x_i, z_i) \big).
\end{equation}
Following~\citet{thomas2015high}, the confidence bounds are:
\begin{equation}
\begin{split}
\textit{CB}(\pi_1,\, &\pi_2 \mid \mathcal{D})
= \frac{7b\ln\big(\frac{2}{1-\epsilon}\big)}{3(|\mathcal{D}|-1)}  \\
& + \frac{1}{|\mathcal{D}|} \sqrt{\frac{ 2 |\mathcal{D}| \ln\big(\frac{2}{1-\epsilon}\big)}{|\mathcal{D}|-1} \sum_{i  \in \mathcal{D}} \big(R_i - \hat{\delta}(\pi_1, \pi_2 \mid \mathcal{D}) \big)^2},
\end{split}
\end{equation}
where $b$ is the maximum possible value for $R_i$.
This results in the lower bound
$
\textit{LCB}(\pi_1, \pi_2 \mid \mathcal{D}) = \hat{\delta}(\pi_1, \pi_2 \mid \mathcal{D}) - \textit{CB}(\pi_1, \pi_2 \mid \mathcal{D}).
$
\ac{GENSPEC} first chooses between the logging policy and $\pi_g$:
\begin{equation}
\pi_{G}(a \mid z)
= 
\begin{cases}
\pi_g(a \mid z),&\text{if } \textit{LCB}(\pi_g', \pi_0 \mid \bounddata) > 0\\
\pi_0(a \mid z),& \text{otherwise},
\end{cases}
\end{equation}
and then per context $z$ whether the policy $\pi_z$ will be activated:
\begin{equation}
\pi_{GS}(a \mid z)
= 
\begin{cases}
\pi_z(a \mid z),&\text{if } \textit{LCB}(\pi_z', \pi_{G}' \mid \bounddata_z) > 0\\
\pi_{G}(a \mid z),& \text{otherwise}. 
\end{cases}
\end{equation}
As such, \ac{GENSPEC} can be applied to the contextual bandit problem for any arbitrary choice of context $z$.

\end{document}